%% file: main.tex
\documentclass[twoside]{article}
\usepackage[accepted]{aistats2025}

\input{math_commands.tex}

\usepackage{hyperref}
\usepackage{url}
\usepackage{graphicx} 
\usepackage{subcaption}
\usepackage{booktabs} %

\usepackage{algpseudocode}
\usepackage{algorithm}
\usepackage{algorithmicx}
\usepackage{caption}

\usepackage{amsthm}

\usepackage[round]{natbib}
\renewcommand{\bibname}{References}
\renewcommand{\bibsection}{\subsubsection*{\bibname}}

\algdef{SE}[DOWHILE]{Do}{doWhile}{\algorithmicdo}[1]{\algorithmicwhile\ #1}
\algnewcommand\algorithmicinput{\textbf{Input:}}
\algnewcommand\pseudoINPUT{\item[\algorithmicinput]}
\algnewcommand\algorithmicoutput{\textbf{Output:}}
\algnewcommand\pseudoOUTPUT{\item[\algorithmicoutput]}
\algnewcommand\algorithmicforeach{\textbf{for each}}
\algdef{S}[FOR]{ForEach}[1]{\algorithmicforeach\ #1\ \algorithmicdo}

\newtheorem{theorem}{Theorem}[section]
\newtheorem{lemma}[theorem]{Lemma}

\begin{document}
\runningauthor{Eduard Tulchinskii, Daria Voronkova, Ilya Trofimov, Evgeny Burnaev, Serguei Barannikov}

\twocolumn[

\aistatstitle{RTD-Lite: Scalable Topological Analysis for Comparing Weighted Graphs in Learning Tasks}

\aistatsauthor{Eduard Tulchinskii \And Daria Voronkova \And Ilya Trofimov }

\aistatsaddress{Skoltech, AI Foundation\\ and Algorithm Lab \And  Skoltech, AIRI \And Skoltech}

\vskip-.2in

\aistatsauthor{Evgeny Burnaev \And Serguei Barannikov}

\aistatsaddress{Skoltech, AIRI \And Skoltech, CNRS}

]

\begin{abstract}
Topological methods for comparing weighted graphs are valuable in various learning tasks but often suffer from computational inefficiency on large datasets. We introduce \mbox{RTD-Lite}, a scalable algorithm that efficiently compares topological features, specifically connectivity or cluster structures at arbitrary scales, of two weighted graphs with one-to-one correspondence between vertices. Using minimal spanning trees in auxiliary graphs, RTD-Lite captures topological discrepancies with \(O(n^2)\) time and memory complexity. This efficiency enables its application in tasks like dimensionality reduction and neural network training. Experiments on synthetic and real-world datasets demonstrate that \mbox{RTD-Lite} effectively identifies topological differences while significantly reducing computation time compared to existing methods. Moreover, integrating RTD-Lite into neural network training as a loss function component enhances the preservation of topological structures in learned representations. Our code is publicly available at \url{https://github.com/ArGintum/RTD-Lite}.
\end{abstract}

\section{INTRODUCTION}

The analysis and comparison of weighted graphs are fundamental tasks in numerous fields such as machine learning, network science, and computational biology. Graph representations are ubiquitous for modeling complex systems, where nodes represent entities, and edges represent relationships or interactions with associated weights. In learning tasks, comparing the structural properties of these graphs can reveal significant insights into the underlying data, such as identifying community structures, detecting anomalies, or understanding the evolution of networks over time.

Topological data analysis (TDA) provides robust tools like persistence barcodes to capture multi-scale topological features of data \citep{barannikov1994framed,zomorodian2001computing,chazal2021introduction}, which are valuable for comparing complex structures like weighted graphs. The Representation Topology Divergence (RTD)~\cite{barannikov2021representation} is one such method that quantifies the topological dissimilarity between two weighted graphs based on comparative cross-barcodes. While RTD provides a rigorous measure of topological differences, it may  become computationally intensive for large-scale graphs due to the high complexity of computing persistence barcodes.

In practical applications involving large datasets, there is a critical need for methods that can efficiently compare the topological features of weighted graphs without compromising on the quality of the insights gained. Existing approaches either fail to capture essential topological characteristics at multiple scales or do not scale well with the size of the data, limiting their applicability in real-world scenarios.

In this paper, we introduce \textbf{RTD-Lite}, a scalable algorithm designed to efficiently compare the topological features of two weighted graphs, focusing particularly on their cluster structures. RTD-Lite leverages the concept of minimal spanning trees (MSTs) within auxiliary graphs constructed from the input graphs to capture topological discrepancies. By avoiding the computational overhead associated with calculating the full persistent homology, RTD-Lite achieves an \(\mathcal{O}(n^2)\) time and memory complexity, where \(n\) is the number of vertices in the graphs.

Key contributions of our work include:
\begin{itemize}
    \item  We propose RTD-Lite as an efficient method to compute a large subset of the RTD features that focuses on zero-dimensional homology (connected components), which is computationally less intensive yet captures essential topological differences between two graphs.

    \item We provide a detailed computational complexity analysis of RTD-Lite, demonstrating its scalability and efficiency compared to existing methods.

    \item We show how RTD-Lite can be  integrated into neural network training as a component of the loss function, promoting the preservation of topological structures in learned representations.

    \item Through extensive experiments on both synthetic and real-world datasets, we demonstrate that RTD-Lite effectively identifies topological differences between graphs. We compare its performance with existing methods, highlighting significant reductions in computation time without sacrificing accuracy.

    \item We explore various applications of RTD-Lite, including dimensionality reduction and analysis of neural network representations, showcasing its practical utility in large-scale learning tasks.
\end{itemize}

By addressing the computational challenges associated with topological graph comparison, RTD-Lite opens new avenues for efficiently analyzing large-scale graphs in various learning tasks. We believe that our contributions will significantly benefit researchers and practitioners working with complex networked data.

\section{RELATED WORK}
\label{sec:related_work}

Comparing weighted graphs is fundamental in domains like machine learning, computational biology and network science.  Traditional methods such as the graph edit distance measure the minimal transformations needed to convert one graph into another \citep{Sanfeliu1983GraphDistance}, but are NP-hard and impractical for large graphs, often focusing on local differences.
Graph kernels, including the Weisfeiler-Lehman graph kernel \citep{Shervashidze2011WeisfeilerLehman} and the shortest-path kernel \citep{Borgwardt2005ShortestPathKernel}, provide scalable comparisons by embedding graphs into feature spaces, capturing structural properties more efficiently \citep{vishwanathan10a}. However, they may struggle with very large graphs and may not capture multi-scale topological features.

Representation Topology Divergence (RTD), introduced in \citet{barannikov2021representation}, measures topological dissimilarities between two weighted graphs with one-to-one correspondence between vertices, based on comparative persistence barcodes. While RTD provides rigorous insights, its computational cost can be relatively high, especially for large graphs. 

Clustering with minimum spanning trees (MSTs) has been explored by \citet{Zahn1971GestaltClusters}, who identified clusters by removing large-weight edges from an MST. \citet{Xu1990ParallelMSTClustering} later enhanced this approach with a fast parallel algorithm, improving scalability for large datasets.

Recent advances have integrated TDA into graph neural networks (GNNs) to enhance graph similarity learning and representation \citet{chen2021topological,wen2024tensor}. New methods also incorporate TDA into community detection in dynamic networks, improving the consistency of cluster structures over time \citep{kong2024learning}. Another direction employs merge trees alongside neural networks to enable topological comparisons of large-scale graphs \citep{qin2024rapid}.

 \cite{luo2021topology} introduced a topology-preserving dimensionality reduction method utilizing a graph autoencoder, while \cite{moor2020topological,trofimov2023learning} proposed  incorporating  additional losses to the autoencoder to maintain the topological structures of the input space in latent representations. 
\cite{kim2020pllay} proposed a differentiable topological layer for general deep learning models based on persistence landscapes. An approach for the differentiation of persistent homology-based functions was introduced by \cite{carriere2021optimizing}, and \cite{leygonie2021framework} established differentiability for maps involving persistence barcodes.

\section{DEFINITION AND TWO ALGORITHMS}
\label{sec:methodology}

Let \( A \) and \( B \) be two connected, undirected, weighted graphs on \( n \) vertices, with a bijection between their vertex sets (so vertex \( i \) in \( A \) corresponds to vertex \( i \) in \( B \)). We assume that the edge weights in both graphs are similarly scaled by aligning their 0.9-quantiles. For an edge \( e \) in \( A \), let \( a_e \) denote its weight and \( b_e \) the weight of the corresponding edge in \( B \) (or plus infinity if the edge is absent). Denote by   \( C \) the auxiliary graph with the same set of vertices and with weights \( c_e = \min(a_e, b_e) \). Everywhere in this work when we talk about the \textit{auxiliary graph} we understand graph \( C \). Thus,  the set of  edges of \( C \) is the union of edge sets of \( A \) and \( B \) with the edge weight equal to the minimum of the weights of the corresponding edges in \( A \) and \( B \).

We propose a method to quantify the discrepancy between multi-scale cluster structures of two graphs, that does it in a similar fashion to RTD but without the necessity to perform persistence barcodes computation. It calculates the set of intervals (barcode) that approximates a $\text{R-Cross-Barcode}_1(A, B)$.
We call it ``RTD-Lite barcode''  and denote it as \textit{RTD-L-barcode}($A, B$); we denote sum of its intervals lengths as RTDL($A, B$).

\textit{RTD-L-barcode}($A, B$) is the collection of intervals $[c_e, a_{t(e)}]$, representing misalignments in multiscale cluster structure of the two weighted graphs. Here, $c_x$ is the level at which a pair of clusters in $C$ merges into one (i.e. the weight of an edge $e$ from minimal spanning tree of $C$). Denote those two clusters (as sets of vertices) as $C_1$ and $C_2$, then the end of the interval ($a_{t(e)}$) will be the minimal level at which some of the vertices from both $C_1$ and $C_2$ belong to the same cluster in $A$ (i.e., in $A$ there is a path between a vertex from $C_1$ and a vertex from $C_2$ such that its maximal edge $t(e)$ has weight of $a_{t(e)}$, and there are no paths with lesser maximal weight edges). Notation $t(e)$ is used to highlight that this `closing' edge depends on edge $e$.

For each edge from minimal spanning tree of $C$, \textit{RTD-L-barcode}($A, B$) contains exactly one interval, $n - 1$ in total, some of them may have zero length and are typically discarded. Figure \ref{fig:clusters_cross_barcodes} presents a direct comparison of RTD-Lite- and R-Cross-Barcodes for a pair of point clouds with $A$ containing one cluster and $B$ containing three clusters shown on figure \ref{fig:clusters2}.

We develop two algorithms to compute RTD-Lite metric. First of them, Algorithm \ref{alg:rtdlite_full}, is the direct implementation of the described procedure. However, often in practical applications instead of the full barcode only the sum of its intervals length is needed. In this case, a faster algorithm can be devised (Algorithm \ref{alg:rtdlite_summonly}). It has better computational time asymptotic but it doesn't reconstruct the intervals, only yields the sum of their lengths. Below, we discuss both algorithms.

\subsection{Direct algorithm}

\begin{algorithm}[th!]
	\caption{Computation of RTD-Lite Barcode} 
    \label{alg:rtdlite_full}

	\begin{algorithmic}
	\pseudoINPUT{$M_A, M_B$ --  weights matrices of  graphs $A$ and $B$ respectively}
	\Require {$MST(\cdot)$ --- function computing Minimal Spanning Tree of a weighted graph, returns list of edges \\ Sort($\cdot$) -- a function sorting list of edges by their weights}
	\pseudoOUTPUT {Multiset of pairs (intervals constructing \textit{RTD-L-barcode}($A, B$))}
	\Procedure{RTD-L-Barcode}{$A, B$}
	\State $M_A, M_B \leftarrow M_A, M_B$ divided by their $0.9$ quantiles
	\State $M_C \leftarrow $ element-wise minimum of $M_A$ and $M_B$
    \State $T_A, T_C \leftarrow MST(A), MST(G_C)$
    \State $T_A, T_C \leftarrow$ Sort($T_A$), Sort($T_C$)
    \State $RTDL\_barcode \leftarrow [\;\empty] $
    \State SubTree $\leftarrow$ empty graph
    \ForEach {$e \in T_C$}
    \State $v_1, v_2 \leftarrow$ vertices of $e$
    \State TemporaryGraph $\leftarrow $ SubTree
    \State $i \leftarrow 0$
    \Repeat
    \State $\tilde{e_i} \leftarrow$ edge $\#i$ from $T_A$
    \State Add $\tilde{e_i}$ to TemporaryGraph
    \State $i \leftarrow i + 1$
    \Until{$v_1\; \text{and}\; v_2$ are connected in TemporaryGraph}
    \State Add ($e; \tilde{e}_{i - 1}$) to $RTDL\_barcode$
    \State Add $e$ to SubTree
    \EndFor
	\State{\textbf{return: }  $RTDL\_barcode$}
	\EndProcedure
	\end{algorithmic}
\end{algorithm}

To compute \textit{RTD-L-barcode}($A, B$) according to the definition, we implement the following procedure (Algorithm \ref{alg:rtdlite_full}).
First, we normalize the weight matrices of both graphs
and construct graph $C$ as edge-wise minimum. We build minimal spanning trees (MSTs) of $A$ and $C$ as two lists of edges sorted by ascension of their weights. 
After that, we iterate over the edges from the MST of graph $C$ and for each edge $e$ (of weight $c_e$) consider a sub-forest containing all previous edges from the MST; then we run over edges from MST of graph $A$ and add them one by one to that sub-forest until there appears a path between ends of edge $e$. Finally, list of intervals $[c_e;a_{t(e)}]$ one for each $e$ from MST of graph $C$ is outputed as \textit{RTD-L-barcode}($A, B$).

\textbf{Proposition 1}. Intervals $[c_e;a_{t(e)}]$ obtained per procedure above are valid (i.e., $c_e \leq a_{t(e)}$).

\textit{Proof: } Suppose the opposite, and $\exists e: c_e > a_{t(e)}$. This means there exists a path between ends of $e$ made from edges $e_1, \ldots, e_l$ from $C$ with weights $\leq c_e$ and edges $e_{l+1}, \ldots, e_k$ from $A$ with weights $\leq a_{t(e)}$ and the largest of them is $t(e)$ with weight of $a_{t(e)}$. $\forall i \in \overline{l + 1, \ldots k}: a_{e_i} \geq c_{e_i}$. Therefore, there is a path in $C$ between ends of $e$ made of edges with weights $\leq a_{t(e)} < c_e$; this contradicts the Cycle Property of MST. 
\hfill$\square$

In our experiments, we operate on full or nearly full graphs thus we implement Prim's algorithm for MST computation. We use \textit{Disjoint Set Union (DSU)} to store sub-forests appear while iterating over edges from MST of $C$. Since DSU does not support erasing of edges (only addition), we make a copy of it before temporarily adding edges from MST of graph $A$.

\paragraph{Computational Complexity.}
The time complexity of weight normalization, and building of graph $C$ is $O(n^2)$. MST construction via Prim's algorithm takes $O(n^2)$, and sorting edges of MST takes $O(n\log n)$ operations. At each iteration of cycle over edges from MST of $C$ we add previous edge to DSU, make a copy of DSU structure (it has $O(n)$ size, therefore $O(n)$ operations) and add to it up to $n - 1$ edges from MST of graph $A$. Later takes $O(n\alpha(n))$ operations where $\alpha(\cdot)$ is the inverse Ackermann function. There are $n - 1$ iterations in this cycle, thus, total time complexity of this algorithm is $O(n^2) + O(n\log n) + (n - 1)O(n) + (n - 1)O(n\alpha(n)) = O(n^2\alpha(n))$.

This algorithm requires $O(n^2)$ additional memory: $O(n^2)$ needed to store graph $C$ weight matrix. Each of the DSU copies requires $O(n)$ memory, but only two of them are needed to be stored at the same time; other steps require no more than linear amount of memory.

\subsection{Simplified computation}

To compute only RTDL$(A, B)$ -- the sum of intervals from the barcode -- a simpler algorithm can be used. 
First, we apply the same preprocessing steps as in previous algorithm: weights normalization and construction of graph $C$. Then we construct MST's for graphs $A$ and $C$ and compute sum of edge weights in them --- $s_A$ and $s_C$ respectively. The required  RTDL$(A, B)$ is $s_A - s_C$.

\begin{algorithm}%
	\caption{RTD-Lite returning only sum of intervals lengths} 
    \label{alg:rtdlite_summonly}

	\begin{algorithmic}%
	\pseudoINPUT{$M_A, M_B$ -- weight matrices of  graphs $A$ and $B$ respectively}
	\Require {$MST(\cdot)$ --- function computing Minimal Spanning Tree of a weighted graph, returns list of edges}
	\pseudoOUTPUT {One real number -- value of RTDL($A, B$)}
	\Procedure{RTD-LiteSumOnly}{$A, B$}
	\State $M_A, M_B \leftarrow M_A, M_B$ divided by their $0.9$ quantiles
	\State $M_C \leftarrow $ element-wise minimum of $M_A$ and $M_B$
    \State $T_A, T_C \leftarrow MST(A), MST(C)$
    \State $s_A \leftarrow \sum_{e \in T_A} w_e$
    \State $s_C \leftarrow \sum_{e \in T_C} w_e$
    \State{\textbf{return: }  $s_A  - s_C$}
	\EndProcedure
	\end{algorithmic}
\end{algorithm}

\textbf{Proposition 2}. Algorithm~\ref{alg:rtdlite_summonly} is correct; i.e., its result is equal to the sum of lengths of intervals yielded by Algorithm~\ref{alg:rtdlite_full} for the same two graphs.

\textit{Proof: } Sum of lengths of intervals in RTD-Lite barcode yielded by Algorithm~\ref{alg:rtdlite_full} is equal to $\sum_{e \in MST(C)}\left(a_{t(e)} - c_e \right) = \sum_{e \in MST(C)}a_{t(e)} - \sum_{e \in MST(C)} c_e = \sum_{e \in MST(C)}a_{t(e)} - s_C$
We need only to show that $s_A = \sum_{e\in MST(A)}a_e = \sum_{e \in MST(C)}a_{t(e)}$. We will do it by proving that each edge from MST of graph $A$ can be $t(e)$ for no more than one $e \in \text{MST}(C)$; bijection between set of all $t(e)$ and MST($A$) will follow from them having the same size.

Suppose the opposite: there are two distinct edges from MST($C$): $(u_1, v_1)$ and $(u_2, v_2)$ such that $t((u_1, v_1)) = t((u_2, v_2)) = (u^0, v^0)$. Let $\prec_A, \prec_C$ be strict orders in which edges 
are listed after being sorted in Algorithm~\ref{alg:rtdlite_full}, and let $(u_1, v_1)~\prec_C~(u_2, v_2)$. Here, as the size of a path we understand the weight of its largest edge.
By its choice, $t((u_1, v_1))$ is the largest edge on the smallest path between $u_1 \text{and} v_1$ made from edges of MST($A$)  and MST($C$) excluding $(u_1, v_1)$.  
That means, there exist two paths: one from $u_1$ to $u^0$ and one from $v_1$ to $v^0$ such that every edge on them is either $\prec_A (u^0, v^0)$ or $\prec_C (u_1, v_1) \prec_C (u_2, v_2)$ (depending from which MST they came). 
Similarly, from choice of $t((u_2, v_2))$ follows presence of paths between $u_2$ and $u^0$ and between $v_2$ and $v^0$ such that every edge on them is either $\prec_A (u^0, v^0)$ or $\prec_C (u_2, v_2)$.

Thus, there is a path between $u_2$ and $u_1$ ($u_2 \rightarrow u^0 \rightarrow u_1$) such that every edge on them is either $\prec_A (u^0, v^0)$ or $\prec_C (u_2, v_2)$ and a similar path between $v_2$ and $v_1$. Since $\prec_C (u_1, v_1) \prec_C (u_2, v_2)$, there is a path between $u_2$ and $v_2$ with the same properties. At least one edge on that path belongs to MST($A$), otherwise together with $(u_2, v_2)$ it will constitute a cycle in MST of graph $C$. That edge is $\prec_A (u^0, v^0)$ but this contradicts the choice of $(u^0, v^0)$ for $t((u_2, v_2))$ in Algorithm~\ref{alg:rtdlite_full} as the smallest edge with such property. 

Therefore, each edge from MST($A$) can correspond (in terms of $t(e)$) to no more than one $e \in \text{MST}(A)$.
\hfill$\square$

\paragraph{Computational Complexity.}
The time complexity of auxiliary steps (weight normalization, and construction of graph $C$) is $O(n^2)$, and Prim's algorithm for MST computation takes $O(n^2)$; total -- $O(n^2)$. Algorithm requires $O(n^2)$ memory -- to store graph $C$.

\subsection{Application to comparison of neural representations}

RTD-Lite can be used to compare two point clouds $P, \tilde{P}$, of the same size $n$ with one-to-one correspondence.
In this case, to estimate the discrepancy, we consider two complete graphs $A, B$ having $n$ vertices with edge weights equal to pairwise distances of $P, \tilde{P}$ respectively.
In particular, $P, \tilde{P}$ can be two neural representations of same objects.
We estimate the topological discrepancy between representations $P,\tilde{P}$ as RTDL$(A,B)$.

\subsection{Gradient Optimization}

 Ends of intervals constituting RTD-Lite barcode are either edge weights from one of the graphs or minimums of weights of two corresponding edges. In both cases, they are (sub-)differentiable by edge weights; therefore for a differentiable function of RTD-Lite barcode (e.g. sum of bars lengths) its (sub-)gradient at edge weights in $A$ and $B$ can be computed. Often edge weights are distances within a point cloud and the (sub-)gradients can be further propagated via the chain rule.
 
\section{RTD-LITE COMPARES MULTI-SCALE CLUSTERS OF TWO GRAPHS}

\label{sec:properties}

In this section, we  compare the connected components of two graphs across various scales.
For any threshold value \(\alpha\), we consider the sub-graphs \(A^{\leq\alpha}\) and \(C^{\leq\alpha}\), which include all edges with weights less than or equal to \(\alpha\) in graphs \(A\) and \(C\), respectively. Notice that the set of edges in \(C^{\leq\alpha}\) is the union of sets of edges in \(A^{\leq\alpha}\) and \(B^{\leq\alpha}\)

The zeroth homology group \(H_0(G)\) over field $k$ of a graph \(G\) is the vector space $k^{\texttt{\#}G}$ where $\texttt{\#}G$ is the set of connected components of $G$. The inclusion of \(A^{\leq\alpha}\) into \(C^{\leq\alpha}\) induces a linear map between their homology groups, denoted as
\[
r_{0,\alpha}: H_0(A^{\leq\alpha}) \rightarrow H_0(C^{\leq\alpha}).
\]
This map sends a connected components of \(A^{\leq\alpha}\) to the  corresponding component in \(C^{\leq\alpha}\).
We have the following exact sequence of linear maps of homology groups:
\begin{equation}
    0 \xrightarrow{} \ker(r_{0,\alpha}) \xrightarrow{\iota} H_0(A^{\leq\alpha}) \xrightarrow{r_{0,\alpha}} H_0(C^{\leq\alpha}) \xrightarrow{} 0,
    \label{eq:exact_sequence}
\end{equation}
where:
 \(0\) denotes the trivial group, \(\ker(r_{0,\alpha})\) is the kernel of \(r_{0,\alpha}\), it is linearly generated by pairs  of connected components in \(A^{\leq\alpha}\) that map to the same component in \(H_0(C^{\leq\alpha})\),  \(\iota\) is the inclusion map.

The sequence~\ref{eq:exact_sequence} is exact, meaning that the image of each map is equal to the kernel of the subsequent map. This exactness captures the precise relationship between the connected components of \(A^{\leq\alpha}\) and \(C^{\leq\alpha}\).
The exact sequence in (\ref{eq:exact_sequence}) provides critical insights into how the topology of graph \(A\) changes when combined with graph \(B\) to form graph \(C\). Specifically: the kernel \(\ker(r_{0,\alpha})\) represents connected components in \(A^{\leq\alpha}\) that are not preserved in \(C^{\leq\alpha}\) due to the influence of \(B\);
the map \(r_{0,\alpha}\) identifies how these components merge or split when transitioning from \(A^{\leq\alpha}\) to \(C^{\leq\alpha}\). The definition of RTD-Lite barcode implies that the dimension of \(\ker(r_{0,\alpha})\) equals the number of intervals in  RTD-Lite barcode that contain $\alpha$, see Appendix.

This analysis allows us to quantify the topological differences between the two graphs at different scales, which is essential for applications that require multiscale connectivity comparisons.

\begin{figure*}[th]
\centering
\begin{subfigure}[t]{\textwidth}
\includegraphics[width=\textwidth]{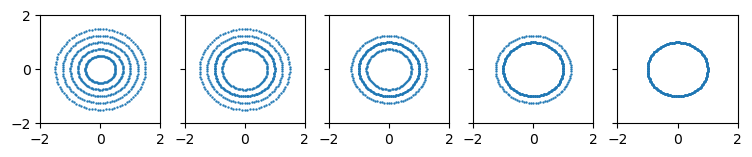}
\caption{Point clouds used in `the `rings'' experiment.}
\end{subfigure}

\begin{subfigure}[t]{\textwidth}
\centering
\includegraphics[width=0.31\textwidth]{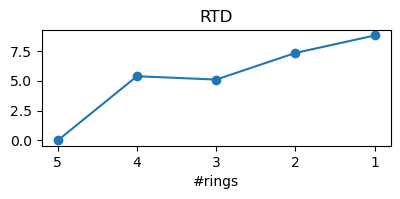}
\includegraphics[width=0.31\textwidth]{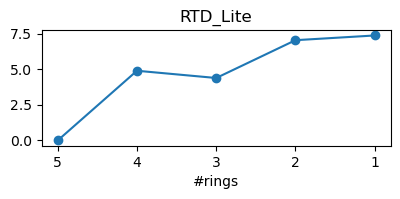}
\includegraphics[width=0.31\textwidth]{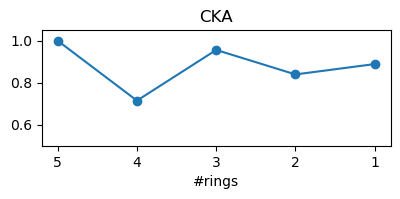}

\caption{Representations' comparison measures. Ideally, the measure should  change monotonically with the increase of topological discrepancy.}
\end{subfigure}

\caption{ RTD-Lite along with RTD perfectly detect cluster structures, while rival measures fail. Five connected components (rings) are compared with 1-5 rings.}

\label{fig:rings2}
\end{figure*}

\section{EXPERIMENTS}
\label{sec:experiments}

\subsection{Comparing point clouds and neural representations}
\label{sec:point_clouds_neural_rep}
\subsubsection{Experiments with synthetic data}

To illustrate the behavior of RTD-Lite for common data comparison scenarios, we perform small-scale experiments on synthetic point clouds: ``Clusters'' and ``Rings''.

\textbf{``Rings'' data}.
We compare synthetic point clouds consisting of a variable number of rings, see Figure~\ref{fig:rings2}, top. The initial point cloud consists of $500$ points uniformly distributed over the unit circle. Then, the points are moved onto circles with radii varying from $0.5$ to $1.5$. Finally, we compare the point cloud having $5$ rings with other ones. Figure \ref{fig:rings2}, bottom, demonstrates 
that both RTD and RTD-Lite almost ideally reflect the change of the topological complexity while CKA fails.

\textbf{``Clusters'' data}. The initial point cloud consists of $300$ points randomly sampled from the $2$-dimensional normal distribution having zero mean (Figure \ref{fig:clusters2}, top). Next, we split it into $2,3,\ldots12$ parts (clusters) and move them to the circle of radius $10$. We compare the initial point cloud (having one cluster) with the split ones by calculating:
RTD \citep{barannikov2021representation},
RTD-Lite and CKA \citep{kornblith2019similarity}, see Figure \ref{fig:clusters2}, top.
Both  RTD-Lite and RTD change monotonically w.r.t. number of clusters, while CKA changes chaotically.
Interestingly, RTD-Lite decreases monotonically, indicating that clusters located on the circle of radius $10$ tend to create a single cluster, like the initial point cloud (the leftmost in Figure \ref{fig:clusters2}, top).

For additional comparisons with SVCCA \citep{raghu2017svcca} and IMD \citep{tsitsulin2019shape} see Appendix \ref{app:aditional_synthetic}.

\begin{figure*}[tb]
\centering
\begin{subfigure}[t]{0.21\textwidth}
\includegraphics[width=\textwidth]{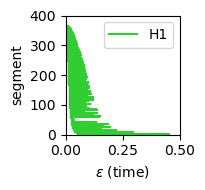}
\caption{R-Cross-Barc.(A, B)}
\end{subfigure}
\begin{subfigure}[t]{0.21\textwidth}
\includegraphics[width=\textwidth]{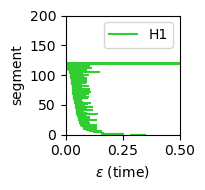}
\caption{R-Cross-Barc.(B, A)}
\end{subfigure}
\begin{subfigure}[t]{0.21\textwidth}
\includegraphics[width=\textwidth]{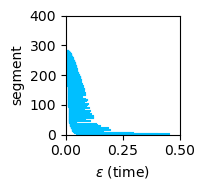}
\caption{RTD-L-Barc.(A, B)}
\end{subfigure}
\begin{subfigure}[t]{0.21\textwidth}
\includegraphics[width=\textwidth]{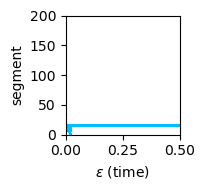}
\caption{RTD-L-Barc.(B, A)}
\end{subfigure}
\vskip-.1in
\caption{Cross-Barcodes for comparison of 1 cluster (A) vs. 3 clusters (B). %
}
\label{fig:clusters_cross_barcodes}
\end{figure*}

\begin{figure*}[t]
\centering
\begin{subfigure}{0.5\textwidth}
\includegraphics[width=0.32\textwidth]{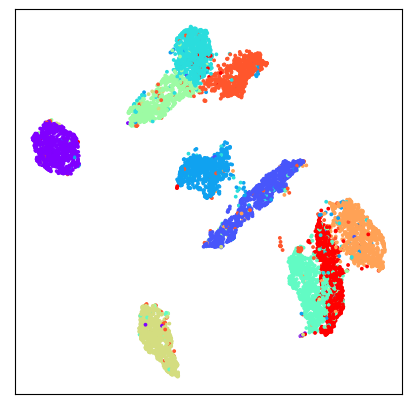}
\includegraphics[width=0.32\textwidth]{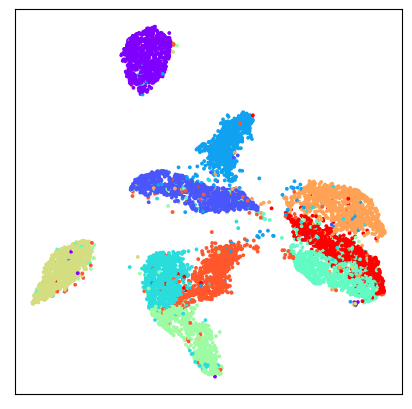}
\includegraphics[width=0.32\textwidth]{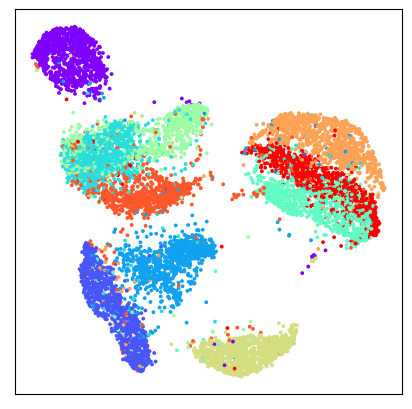}
\caption{2D representations of MNIST.}
\label{fig:umap_mnist_3vars}
\end{subfigure}
\begin{subfigure}{0.142\textwidth}
\vskip-1in
\includegraphics[width=\textwidth]{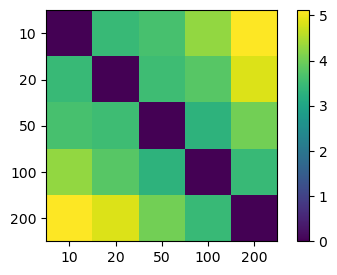}
\caption{RTD}
\end{subfigure}
\begin{subfigure}{0.15\textwidth}
\includegraphics[width=\textwidth]{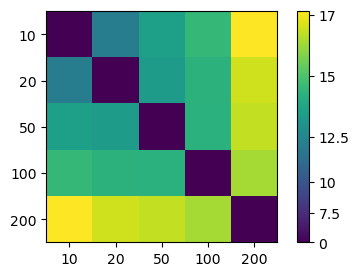}
\caption{RTD-Lite}
\end{subfigure}
\begin{subfigure}{0.147\textwidth}
\includegraphics[width=\textwidth]{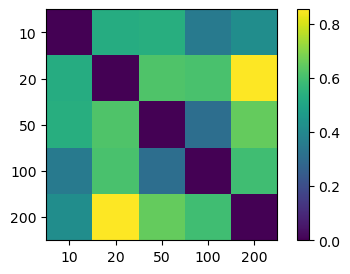}
\caption{1-CKA}
\label{fig:umap_mnist_cka}
\end{subfigure}

\caption{Comparing representations of MNIST by UMAP with varying n\_neighbors. Left: representations with n\_neighbors = 10, 50, 200. Right: comparison of representations by metrics.}
\label{fig:umap_mnist_cka_rtd}
\end{figure*}

\subsubsection{Visualization of RTD-Lite-Barcodes}

RTD-Lite also creates a list of intervals which is called RTD-Lite Barcodes (see Algorithm \ref{alg:rtdlite_full}) depicting a multi-scale differences in topological structures. R-Cross-Barcodes from \cite{barannikov2021representation} and RTD-Lite Barcodes are depicted in Figure \ref{fig:clusters_cross_barcodes}. The barcodes have some similarities, see more examples in Appendix \ref{app:rtd-lite-barcodes}.

\subsubsection{Comparing representations from UMAP}
\label{sec:umap_mnist}

In this experiment, we verify whether RTD-Lite is able to capture the growing topological dissimilarity of representation spaces of real data. For this purpose, we apply UMAP \citep{mcinnes2018umap}, the state-of-the-art dimensionality reduction method, to get 2D representations of the MNIST dataset. In UMAP, we vary the number of neighbors in the range $(10, 20, 50, 100, 200)$, see Figure \ref{fig:umap_mnist_3vars}. The number of neighbors in UMAP affects the topological structure: for low values, the algorithm focuses on the local structure and clusters are crisp; for high values, the algorithm pays more attention to the global structure, and clusters overlap often. Then, we perform the pairwise comparison of all the variants of 2D representations by RTD, RTD-Lite and CKA, see Figure \ref{fig:umap_mnist_cka_rtd}.
CKA reveals a chaotic pattern w.r.t. number of neighbors. In contrast, both RTD and RTD-Lite have monotonic increasing behavior which coincides with the growing dissimilarity in UMAP representations.

\subsubsection{Comparing Representations of DNN layers}
\label{diff_init}

\begin{figure*}[t]
\centering
\begin{subfigure}{0.23\textwidth}
\includegraphics[width=\textwidth]{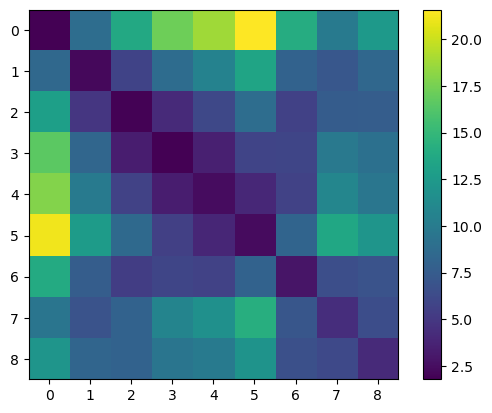}
\caption{RTD}
\end{subfigure}
\begin{subfigure}{0.23\textwidth}
\includegraphics[width=\textwidth]{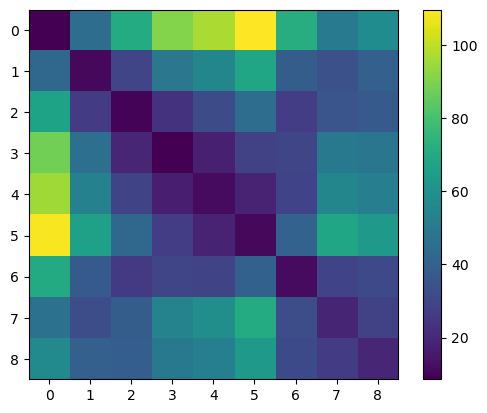}
\caption{RTD-Lite}
\end{subfigure}
\begin{subfigure}{0.23\textwidth}
\includegraphics[width=\textwidth]{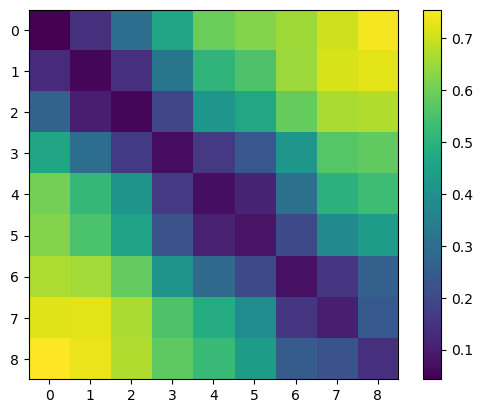}
\caption{1-CKA}
\end{subfigure}
\begin{subfigure}{0.25\textwidth}
\includegraphics[width=\textwidth]{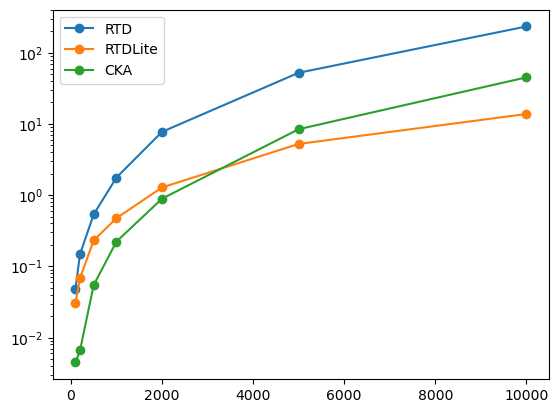}
\caption{Running times of RTD, RTD-Lite, and CKA w.r.t number of samples. The vertical axis is log-scaled.}
\label{fig:timing}
\end{subfigure}
\caption{Comparing layers from All-CNN networks trained on CIFAR10 with different seeds.}
\label{fig:cnn_layers}

\end{figure*}

\begin{figure*}[tbp]
\centering
\begin{subfigure}{0.17\textwidth}
\includegraphics[width=\textwidth]{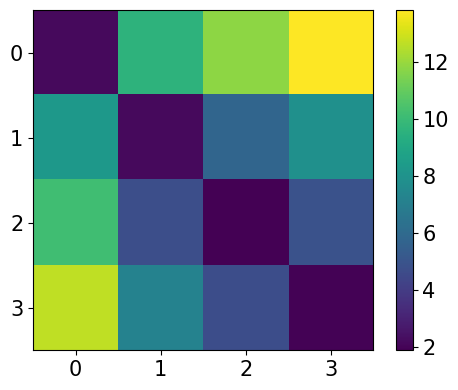}
\caption{GCN}
\label{fig:gnn_sanity_a}
\end{subfigure}
\begin{subfigure}{0.17\textwidth}
\includegraphics[width=\textwidth]{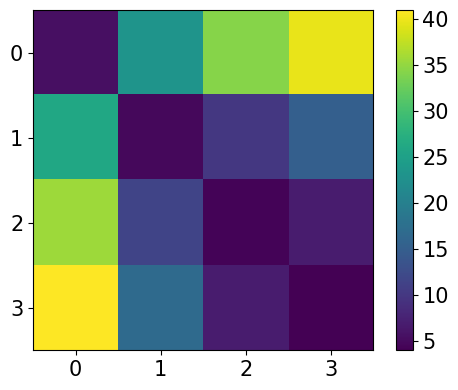}
\caption{SAGE}
\label{fig:gnn_sanity_b}
\end{subfigure}
\begin{subfigure}{0.17\textwidth}
\includegraphics[width=\textwidth]{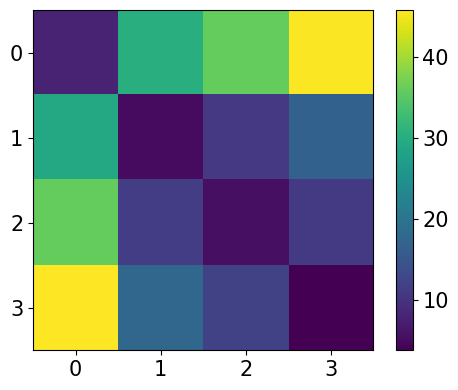}
\caption{GAT}
\label{fig:gnn_sanity_c}
\end{subfigure}
\begin{subfigure}{0.17\textwidth}
\includegraphics[width=\textwidth]{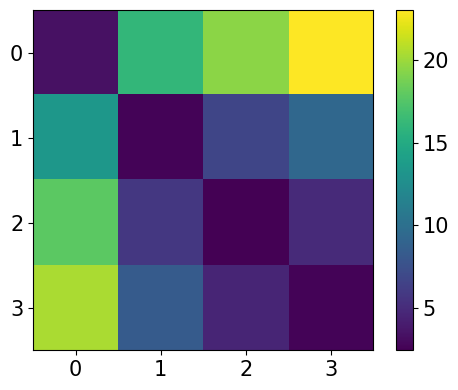}
\caption{CGCN}
\label{fig:gnn_sanity_d}
\end{subfigure}
\begin{subfigure}{0.2\textwidth}
\includegraphics[width=\textwidth]{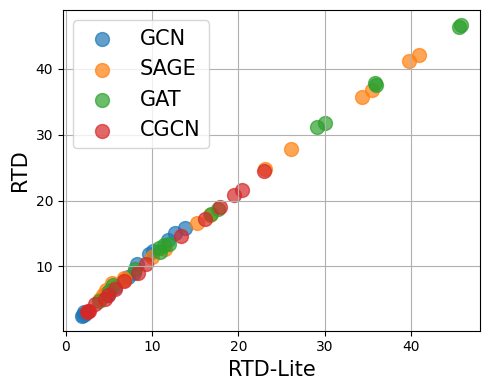}
\caption{RTD-Lite vs. RTD}
\label{fig:gnn_sanity_e}
\end{subfigure}
\caption{RTD-Lite scores between layers from GNNs trained with different seeds.}
\label{fig:gnn_sanity_1}
\end{figure*}
\vskip-.1in

We evaluate the proposed RTD-Lite with sanity tests: in pairwise comparisons of layers' representations from networks trained with different initializations, the corresponding layers are expected to be the closest ones. To account for versatile structure of the underlying data, we perform this experiment for both CNN and GNN architectures.

\textbf{CNN architecture.} We train All-CNN networks \cite{springenberg2014striving} on CIFAR-$10$ with $10$ different seeds. Figure \ref{fig:cnn_layers} depicts averaged comparisons of representations from all convolutional layers (after batch normalization). RTD-Lite provides an accurate estimate of the RTD behavior pattern. We also carried out a classification experiment where for each layer the predicted corresponding layer was selected by metrics under analysis. The classification accuracy obtained: RTD - 98\%, RTD-Lite - 98\%, CKA - 96\%. 
The calculation of RTD-Lite was $\sim$ 6 times faster than RTD.

\textbf{GNN architecture.} We strictly follow the setup from \cite{ebadulla2024normalized} and train four GNN architectures — GCN \citep{kipf2016semi}, GraphSAGE \citep{hamilton2017inductive}, GAT~\mbox{\citep{velickovic2017graph}} and ClusterGCN \citep{chiang2019cluster} — to solve Node Classification task on the Amazon Computers dataset \citep{shchur2018pitfalls}. Figures \ref{fig:gnn_sanity_a}-\ref{fig:gnn_sanity_d} demonstrates that RTD-Lite has lowest values for the corresponding layers among all pairwise comparisons for all the architectures. By comparing RTD and RTD-Lite scores in Figure \ref{fig:gnn_sanity_e}, we reveal that both metrics are well aligned.

\subsection{Running times and computational complexity}
To compare two representations of $n$ objects having dimensionalities $d_1$, $d_2$, the methods under analysis have the following computational complexity:
CKA - $O(n^3 + n^2 (d_1 + d_2))$, 
\mbox{RTD-Lite} - $O(n^2 (\alpha(n) + d_1 + d_2))$.
For computation of RTD, one should calculate pairwise distances having cost $O(n^2 (d_1 + d_2))$.
However, the computation of RTD is dominated by the persistence barcode evaluation, which is at worst cubic in the number of simplices involved. The later one increases at least exponentially with $n$. In practice, the computation is much faster since the boundary matrix is typically sparse for real datasets.

Running times of RTD, RTD-Lite, and CKA w.r.t. number of samples $n$ are depicted in Figure \ref{fig:timing}. We used embeddings of CIFAR10 from a CNN for testing. RTD-Lite is significantly faster than RTD.
For small $n$, CKA is fast probably because of a multicore implementation of matrix multiplication in \texttt{numpy}.
For large $n$, RTD-Lite is the fastest method.

Additionally, we carried out an experiment for comparison of layers’ representations for a larger data of size 50k (full CIFAR10 train dataset) corresponding to graphs having 50k vertices. Running times are the following:
RTD-Lite: 3.75 minutes, RTD: out of GPU memory, CKA: 36 minutes. Standard RTD metric is not applicable for such large data, while RTD-Lite is ~10 times faster than CKA.

\subsection{Analysis of gradient optimization}
\label{sec:dim_reduction}

Here we show that our method can be used not only as static metric but also as a loss function component during training of neural networks. We illustrate this on task of dimensionality reduction. Here we follow the setup proposed in \cite{trofimov2023learning} and implement a deep autoencoder in which loss function has integrated component of RTD-Lite between original high-dimensional data and their low-dimensional representations. Therefore, given the data $X = \lbrace x_i\rbrace_{i=1}^{n}, x_i \in \mathrm{R}^d$ and its latent representation $Z = \lbrace z_i\rbrace_{i=1}^{n}, z_i \in \mathrm{R}^d$, the training loss function will be $\frac{1}{n}\lvert\rvert X - X_{rec}\lvert\rvert_2^2 + \frac{\lambda}{2n}\left( \text{RTDL}(X, Z) + \text{RTDL}(Z, X)\right)$, where $\lambda$ is a hyperparameter. This loss is computed on mini-batches.

We compare the proposed RTD-Lite AE againts the  RTD-AE \cite{trofimov2023learning}, as well as vanilla autoencoder (AE), and TopoAE \cite{moor2020topological}. Also, for reference, we provide results of applying UMAP \cite{damrich2021umap} to the same data. Autoencoder models implement the same architecture and differ only in the training loss. The complete descriptions of datasets and hyperparameters used are provided in Appendix. We evaluate these methods by four metrics comparing original data and their latent representations. They are: (1) linear correlation of pairwise distances, (2) Wasserstein distance between $H_0$ persistence barcodes \cite{chazal2021introduction}, (3) Triplet distance ranking Accuracy \cite{wang2021understanding}, and finally (4) RTD. All of these metrics are tailored at measuring how good the manifold's global structure and topology are preserved within latent representations; however, they work on different principles and capture different aspects of the matter; therefore, we report all four of them.

\textbf{Synthetic data}.
First, we used our method at a synthetic dataset ``Spheres''. It consists of 10,000 points randomly sampled from surfaces of eleven 100D spheres lying in $\mathrm{R}^{101}$; none of them intersect, and one contains all others inside.

Results are presented in Table \ref{tbl:measure_synthetic3d}. From it we can see that
RTD-Lite autoencoder performs slightly worse than RTD-AE or TopoAE but overall is still on somewhat equal level with them. RTD-AE achieves the best quality by almost all metrics. Training of autoencoder with RTD-Lite loss, however, is much faster than of one with RTD loss and unlike later it does not require the two-stage procedure when autoencoder is trained with reconstruction loss only for first several epochs.

\addtolength{\tabcolsep}{-2.5pt}
\begin{table}
  \caption{Quality of data manifold global structure preservation at projection Spheres dataset.}
  \label{tbl:measure_synthetic3d}
  \centering
  \begin{tabular}{lcccc}
    \toprule
    & \multicolumn{4}{c}{Quality measure}           \\
    \cmidrule(r){2-5}
    Method   & L. C. & W. D. $H_{0}$ & T. A. & RTD \\
    \midrule
    \multicolumn{5}{c}{\textbf{into 2D space}} \\
    UMAP & 0.021 & {44.90 $\pm$ 1.8} & 0.54 $\pm$ 0.01 & 42.60 $\pm$ 1.8 \\
    AE & 0.311 & {46.07 $\pm$ 1.5}  & 0.41 $\pm$ 0.01 & 41.02 $\pm$ 1.4 \\
    TopoAE & {0.495} & \underline{43.92 $\pm$ 2.5} & {0.54 $\pm$ 0.02} & \underline{39.69 $\pm$ 1.4} \\
    RTD & \underline{0.626} & {45.29 $\pm$ 2.2} &\underline{0.68 $\pm$ 0.02}  & \underline{39.60 $\pm$ 1.9}\\
    RTD-L& {0.570} & {45.72 $\pm$ 2.0} & {0.64 $\pm$ 0.02} & \underline{40.00 $\pm$ 1.7} \\
    \midrule
    \multicolumn{5}{c}{\textbf{into 3D space} } \\
    UMAP & 0.041 & 45.56 $\pm$ 2.2 & 0.54 $\pm$ 0.01 & 41.79 $\pm$ 2.2 \\
    AE & 0.376 & {46.06 $\pm$ 2.3} & 0.41 $\pm$ 0.02 & 41.05 $\pm$ 2.2 \\
    TopoAE & {0.622} & \underline{40.41 $\pm$ 2.7} & 0.71 $\pm$ 0.01 & 34.74 $\pm$ 1.9 \\
    RTD & \underline{0.687} & {41.39 $\pm$ 2.3} & \underline{0.74 $\pm$ 0.02} & \underline{33.74 $\pm$ 1.6} \\
    RTD-L & 0.615 & {41.76 $\pm$ 1.8} & 0.66 $\pm$ 0.02 & 35.80 $\pm$ 1.5 \\
    \bottomrule
  \end{tabular}
\end{table}
\addtolength{\tabcolsep}{2pt}
\vspace{-1mm}

\addtolength{\tabcolsep}{-2.5pt}
\begin{table}
\caption{Global structure preservation quality during dimensionality reduction of real-life datasets.}
\centering
\label{tbl:real16dim}
\begin{tabular}{lcccc}
    \toprule
    & \multicolumn{4}{c}{Quality measure}           \\
    \cmidrule(r){2-5}
    Method   & L. C. & W. D. $H_{0}$ &  T. A. & RTD \\
    \midrule
    \multicolumn{5}{c}{\textbf{F-MNIST into 16D space}} \\
    UMAP & 0.602 & 592.0 $\pm$ 3.9 & 0.741 $\pm$ 0.02 & 12.31 $\pm$ 0.4\\
    AE & 0.879 & 320.5 $\pm$ 1.9 & 0.850 $\pm$ 0.00 & 5.52 $\pm$ 0.2  \\
    TopoAE & 0.905 & 190.7 $\pm$ 1.2 & 0.867 $\pm$ 0.01 & 3.69 $\pm$ 0.2 \\
    RTD & \underline{0.960} &\underline {181.2 $\pm$ 0.8} & \underline{0.907 $\pm$ 0.00} & \underline{3.01 $\pm$ 0.1}\\
    RTD-L & {0.933} & \underline{181.7 $\pm$ 1.4} & \underline{0.910 $\pm$ 0.01} & {3.75 $\pm$ 0.2}\\
    \midrule
    \multicolumn{5}{c}{\textbf{COIL-20 into 16D space}} \\
   UMAP & 0.301 & 274.7 $\pm$ 0.0 &  0.574 $\pm$ 0.01 & 15.99 $\pm$ 0.5\\
    AE   & 0.834 & 183.6 $\pm$ 0.0 &  0.809 $\pm$ 0.01 & 8.35 $\pm$ 0.2\\
    TopoAE & 0.910 & 148.0 $\pm$ 0.0 & 0.822 $\pm$ 0.02 & 6.90 $\pm$ 0.2\\
    RTD & {0.944} & {88.9 $\pm$ 0.0} & {0.892 $\pm$ 0.01} & {5.78 $\pm$ 0.1}  \\
    RTD-L & \underline{0.949} & \underline{77.1 $\pm$ 0.0} & \underline{0.898 $\pm$ 0.01} & \underline{5.35 $\pm$ 0.2}\\
   \bottomrule
\end{tabular}
\end{table}
\addtolength{\tabcolsep}{2pt}
\vspace{-2mm}
\textbf{Real-life datasets}.
We performed similar style experiments with two real-world datasets: F-MNIST \citep{xiao2017online}, and COIL-20 \citep{nene1996columbia}. Unlike \textit{Spheres}, these datasets doesn't have such clear multi-dimensional internal structures and are much more noisy. These datasets have much higher number of parameters, and so for them we choose latent dimension of size 16. Table \ref{tbl:real16dim} presents the results. In this situation, RTD-Lite AE performs at level of \mbox{RTD-AE} and noticeably better than other methods.

\section{CONCLUSION}
\label{sec:conclusion}

Our method, RTD-Lite, streamlines topological feature computation for graphs having one-to-one correspondence between nodes by focusing on comparing multi-scales cluster structures via MSTs on auxiliary graphs, achieving $\mathcal{O}(n^2)$ complexity. This reduces the computational load significantly w.r.t standard RTD metric.
RTD-Lite works well as a tool for comparing neural representations of CNNs and GNNs. Moreover, RTD-Lite integrates efficiently as a loss function in neural network training for dimensionality reduction, preserving both global and local data structures without the overhead of full persistence barcode calculations.

\subsubsection*{Acknowledgments}
Research was partially supported by the funding of Skoltech Applied AI center.

\bibliography{References}

\clearpage

\section*{Checklist}

 \begin{enumerate}

 \item For all models and algorithms presented, check if you include:
 \begin{enumerate}
   \item A clear description of the mathematical setting, assumptions, algorithm, and/or model. [Yes, Sections \ref{sec:methodology}, \ref{sec:properties}]
   \item An analysis of the properties and complexity (time, space, sample size) of any algorithm. [Yes, Sections \ref{sec:methodology}, \ref{sec:properties}]
   \item (Optional) Anonymized source code, with specification of all dependencies, including external libraries. [Yes]
 \end{enumerate}

 \item For any theoretical claim, check if you include:
 \begin{enumerate}
   \item Statements of the full set of assumptions of all theoretical results. [Yes, Sections \ref{sec:methodology}, \ref{sec:properties}, Appendix]
   \item Complete proofs of all theoretical results. [Yes, Appendix]
   \item Clear explanations of any assumptions. [Yes, Sections \ref{sec:methodology}, \ref{sec:properties}]     
 \end{enumerate}

 \item For all figures and tables that present empirical results, check if you include:
 \begin{enumerate}
   \item The code, data, and instructions needed to reproduce the main experimental results (either in the supplemental material or as a URL). [Yes]
   \item All the training details (e.g., data splits, hyperparameters, how they were chosen). [Yes, Appendix]
         \item A clear definition of the specific measure or statistics and error bars (e.g., with respect to the random seed after running experiments multiple times). [Yes]
         \item A description of the computing infrastructure used. (e.g., type of GPUs, internal cluster, or cloud provider). [Yes]
 \end{enumerate}

 \item If you are using existing assets (e.g., code, data, models) or curating/releasing new assets, check if you include:
 \begin{enumerate}
   \item Citations of the creator If your work uses existing assets. [Yes]
   \item The license information of the assets, if applicable. [Yes, Appendix~\ref{sec:licenses}]
   \item New assets either in the supplemental material or as a URL, if applicable. [Yes]
   \item Information about consent from data providers/curators. [Not Applicable]
   \item Discussion of sensible content if applicable, e.g., personally identifiable information or offensive content. [Not Applicable]
 \end{enumerate}

 \item If you used crowdsourcing or conducted research with human subjects, check if you include:
 \begin{enumerate}
   \item The full text of instructions given to participants and screenshots. [Not Applicable]
   \item Descriptions of potential participant risks, with links to Institutional Review Board (IRB) approvals if applicable. [Not Applicable]
   \item The estimated hourly wage paid to participants and the total amount spent on participant compensation. [Not Applicable]
 \end{enumerate}

 \end{enumerate}

\clearpage

\newpage

\onecolumn
\appendix

\section{Proof of the exact sequence}

We provide a brief proof of the exactness of the short sequence in Equation \ref{eq:exact_sequence}.
\paragraph{Exactness at \(\ker(r_{0,\alpha})\):} The kernel of inclusion map  \(\iota\) is trivial and coincides with the image of the map from 0. %
\paragraph{Exactness at \(H_0(A^{\leq\alpha})\):} The image of \(\iota\) is \(\ker(r_{0,\alpha})\), and the kernel of \(r_{0,\alpha}\) is exactly \(\ker(r_{0,\alpha})\), satisfying the condition \(\mathrm{Im}(\iota) = \ker(r_{0,\alpha})\).
\paragraph{Exactness at \(H_0(C^{\leq\alpha})\):} Since \(r_{0,\alpha}\) is surjective onto its image, because each connected component of \(C^{\leq\alpha}\) contains at least one vertex from \(A^{\leq\alpha}\), and the sequence ends with \(0\), the exactness condition at \(H_0(C^{\leq\alpha})\) is satisfied.

\section{On some intuition behind RTD-Lite}
\label{sec:rtdl_intuition}

Given a finite weighted graph $G$, consider a sequence of its subgraphs $G_\alpha$, such that $G_\alpha$ contains all vertices of $G$ and all edges with weights $\leq \alpha$. Let this sequence contain all unique $G_\alpha$ for all meaningful $\alpha$ (it is finite because there is a finite number of different edge weights). 
With each $G_\alpha$ we associate a simplicial complex, containing a $k$-dimensional simplex for each $k + 1$-element clique in $G_\alpha$. There is a natural mapping $G_\alpha \rightarrow G_\beta$ for $\alpha \leq \beta$, which can be extended to the elements of the simplicial complexes associated with $G_\alpha$ and $G_\beta$. Thus, we obtain a \textit{simplicial filtration}.

Essentially, $0-$dimensional topological features are  connected components in the graph, $1-$dimensional are (equivalence classes) of loops. $2-$ and higher ($k-$) dimensional features
correspond to $k - 1$-dimensional voids.

 Persistence barcode (\cite{barannikov2021canonical}) describes the lifespans of topological features. The persistence barcode is a multiset of intervals $[b_i, d_i]$ corresponding to basic topological features where $b_i$ is the threshold at which the topological feature $i$ first appears (birth moment) in the associated filtration and $d_i$ is the threshold at which it disappears (death moment). One can limit oneself to only the features formed by $0-, 1-, \ldots k-$th simplices, thus, obtaining $0-,1-, \ldots k-$ \textit{dimensional barcodes}.

A topological discrepancy RTD-Lite feature appears at threshold $a = b_e$, when in the union of edges of $A^{\leq e}$ and $B^{\leq e}$, two vertex sets $C_1$ and $C_2$ disjoint at smaller thresholds, are joined into one connected component by the edge $e$ from $B^{\leq e}$. This interval is closed at threshold $\alpha = a_{e\prime}$ when the two vertex sets $C_1$ and $C_2$ are joined into one connected component by the edge $e^\prime$ in the graph of $A^{\leq \alpha}$.

Essentially, R-Cross-Barcode, comparatively to RTD-Lite Barcode, contains in addition the features that describe non-trivial 1-cycles appearing in one graph at some scale but not yet appearing in another graph.

\section{Details on the experiment with ``clusters''}
\label{app:rtd-lite-barcodes}

\begin{figure*}[tbp]
\centering
\begin{subfigure}[t]{\textwidth}
\includegraphics[width=\textwidth]{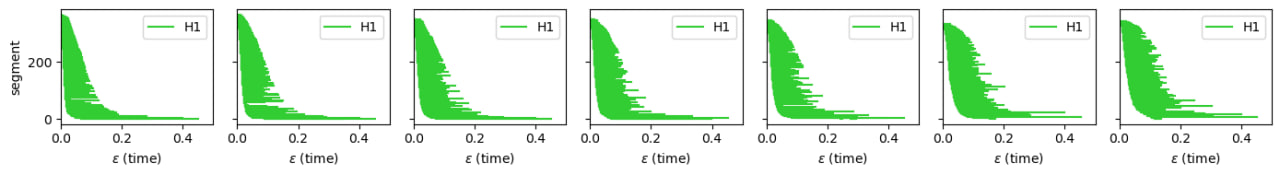}
\caption{R-Cross-Barcode(A,B).}
\end{subfigure}

\begin{subfigure}[t]{\textwidth}
\centering
\includegraphics[width=\textwidth]{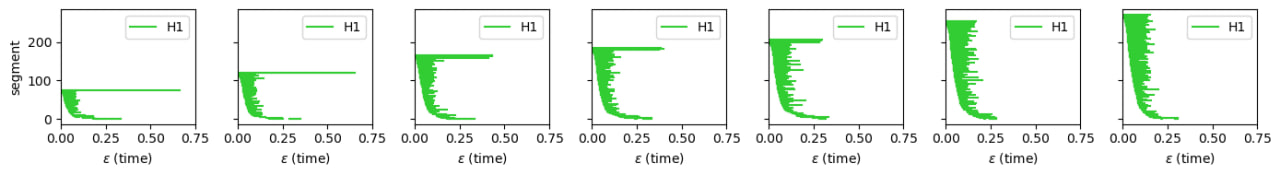}
\caption{R-Cross-Barcode(B,A).}
\includegraphics[width=\textwidth]{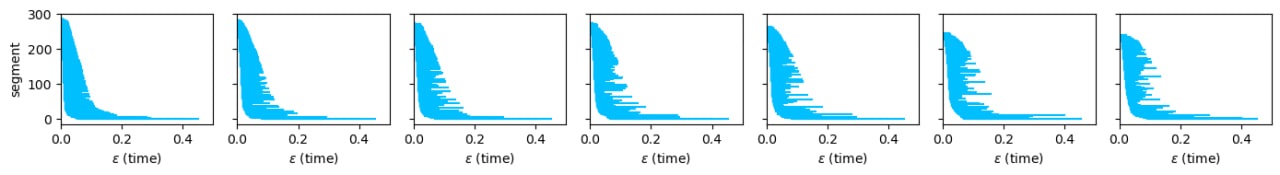}
\caption{RTD-Lite Barcode(A,B).}

\includegraphics[width=\textwidth]{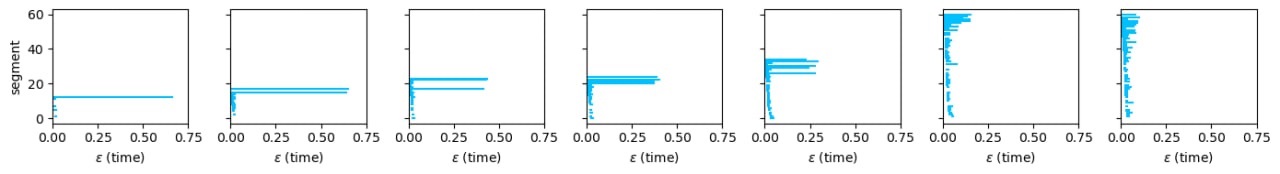}
\caption{RTD-Lite Barcode(B,A).}
\end{subfigure}

\caption{Cross-Barcodes for comparison of one cluster vs. 2-12
clusters.}

\label{fig:clusters_all_barcodes}
\end{figure*}

\begin{figure*}[tb]
\centering
\begin{subfigure}[t]{\textwidth}
\includegraphics[width=\textwidth]{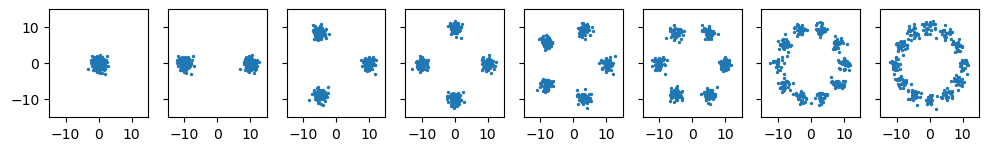}
\caption{Point clouds used in the ``clusters'' experiment.}
\end{subfigure}

\begin{subfigure}[t]{\textwidth}
\centering
\includegraphics[width=0.31\textwidth]{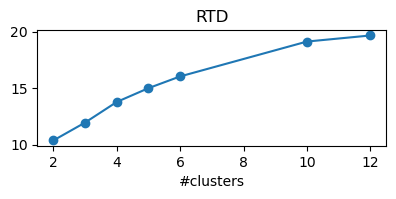}
\includegraphics[width=0.31\textwidth]{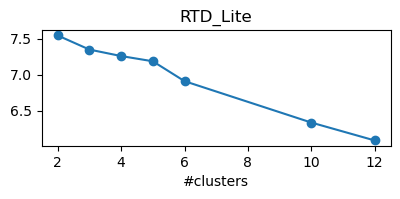}
\includegraphics[width=0.31\textwidth]{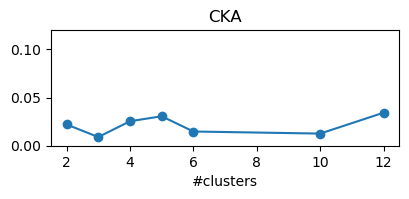}

\caption{Representations' comparison measures. Ideally, the measure should  change monotonically with the increase of topological discrepancy.}
\end{subfigure}

\caption{Comparison of point clouds via RTD, RTD-Lite, CKA. One cluster is compared to 2-12 clusters.}

\label{fig:clusters2}
\end{figure*}

Figure \ref{fig:clusters2} presents point clouds under comparison: one cluster vs. 2-12 clusters.
Figure \ref{fig:clusters_all_barcodes} shows R-Cross-Barcodes and RTD-Lite barcodes for the comparisons which have similarities.

\section{Additional results for synthetic datasets}
\label{app:aditional_synthetic}

\cite{barannikov2021representation} carried out experiments with the same datasets ``rings'' and ``clusters'' and compared RTD with CKA, SVCCA, IMD. As a quality measure, \cite{barannikov2021representation} used Kendall-tau correlation between a metric and a number of rings and clusters respectively. We have calculated the Kendall-tau correlation for RTD-Lite and merged with results of \cite{barannikov2021representation}. 

For ``rings'', the Kendall-tau rank correlations of the measures with a number of rings are:

RTD-Lite: 0.8, RTD: 0.8, CKA: -0.2, IMD: 0.8, SVCCA: -0.2.

For ``clusters'', the Kendall-tau rank correlations of the measures with a number of clusters are:

RTD-Lite: -1.0, RTD: 1.0, CKA: 0.23, IMD: 0.43, SVCCA: 0.14.

In both of these cases, RTD and RTD-Lite show the highest absolute Kendall-tau correlation, meaning that RTD and RTD-Lite correctly capture changes of topological structures. However, for the experiment with “clusters” the dependency between RTD-Lite and the number of clusters is reversed. We would like to emphasize that we do not expect that RTD-Lite will be more sensitive than RTD. Its main benefit is calculation speed.

\section{Analysis of RTD-Lite for GNN representations}

In line with the recent research \citet{ebadulla2024normalized}, we provide analysis of RTD-Lite for measuring dissimilarity in GNN representations as additional verification for a domain with an inherent complex structure.

\subsection{Convergence Analysis}
\label{app:gnn_convergence}

\begin{figure*}[t]
\centering
\includegraphics[width=\textwidth]{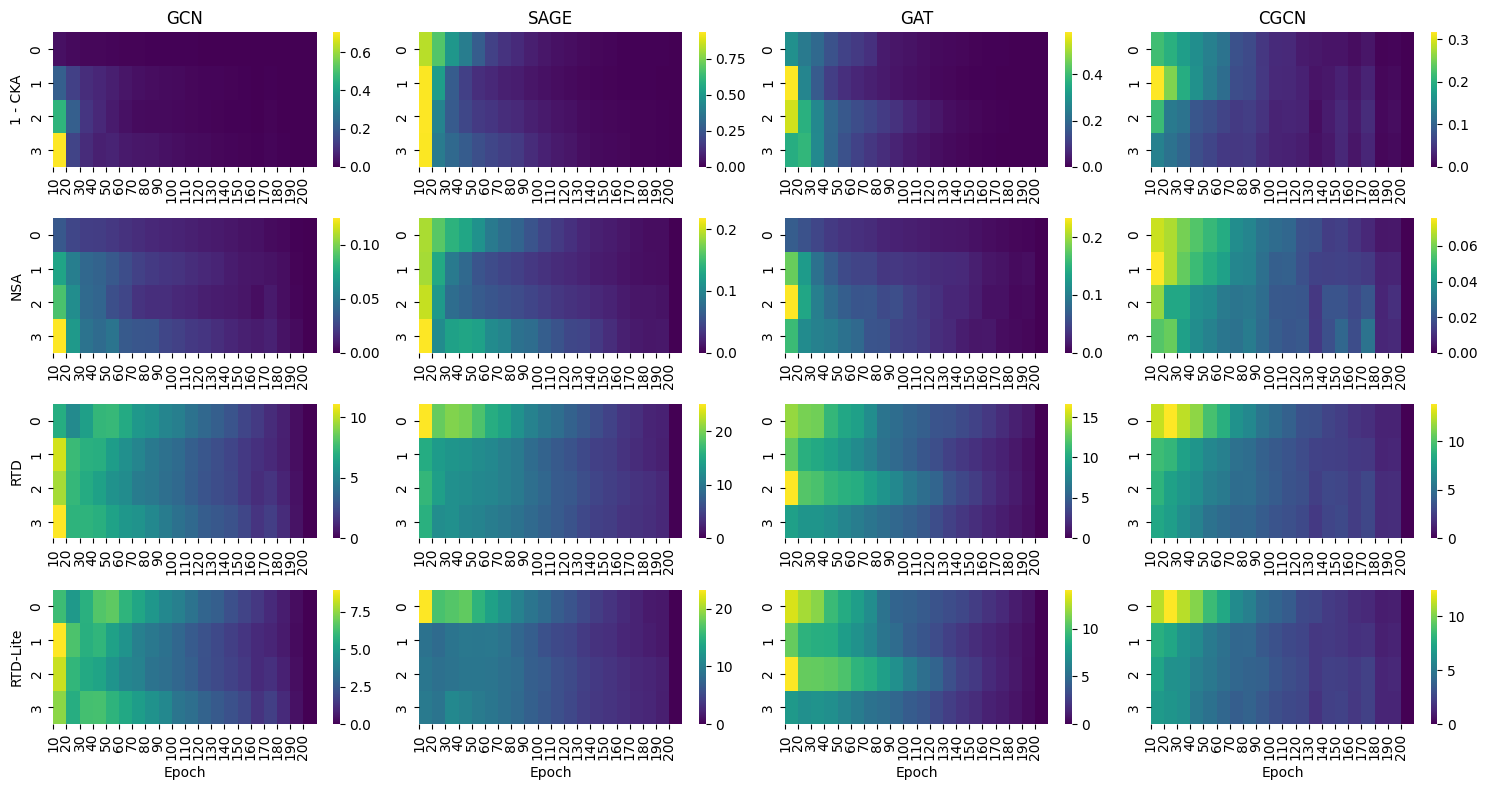}
\caption{Evolution of dissimilarity in GNNs' representations at an intermediate and final training epochs as measured by CKA, NSA, RTD, RTD-Lite.}
\label{fig:gnn_converg_1}
\end{figure*}

\begin{figure*}[tbp]
\centering
\begin{subfigure}{0.24\textwidth}
\includegraphics[width=\textwidth]{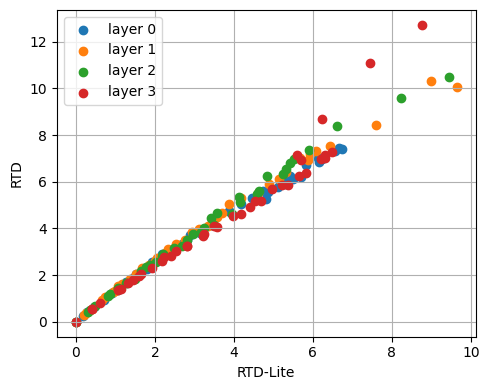}
\caption{GCN}
\label{fig:gnn_converg_2_gcn}
\end{subfigure}
\begin{subfigure}{0.24\textwidth}
\includegraphics[width=\textwidth]{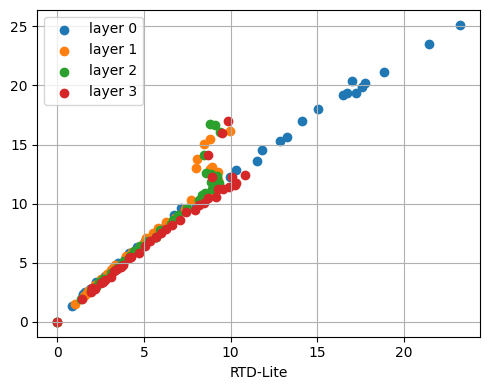}
\caption{SAGE}
\label{fig:gnn_converg_2_sage}
\end{subfigure}
\begin{subfigure}{0.24\textwidth}
\includegraphics[width=\textwidth]{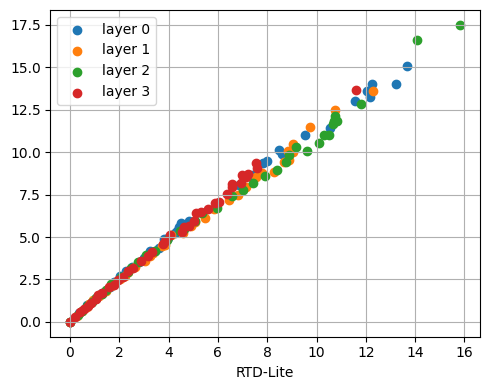}
\caption{GAT}
\label{fig:gnn_converg_2_gat}
\end{subfigure}
\begin{subfigure}{0.24\textwidth}
\includegraphics[width=\textwidth]{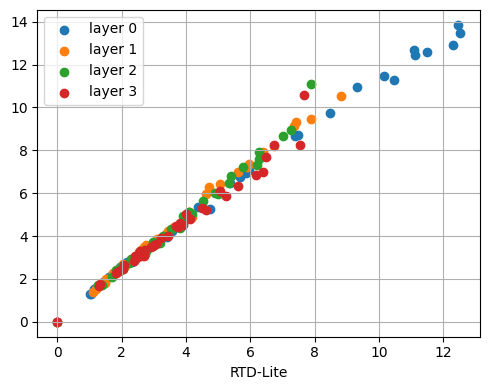}
\caption{CGCN}
\label{fig:gnn_converg_2_cgcn}
\end{subfigure}
\caption{RTD-Lite vs. RTD in measuring dissimilarity between GNN layers' representations at an intermediate and the final training epochs.}
\label{fig:gnn_converg_2}
\end{figure*}

In this experiment, we inspect how layers' representations from a GNN evolve during training as measured by the metrics under analysis: CKA, NSA \citep{ebadulla2024normalized}, RTD, RTD-Lite. We strictly follow the setup from \citet{ebadulla2024normalized} and, for each layer, we compare its representations from an intermediate epoch with its representations from the final epoch of training (Figure \ref{fig:gnn_converg_1}). In general, dissimilarity in representations decreases throughout training as measured by RTD-Lite. Although there are several cases of non-monotonic behaviour in the beginning of training, they are mostly revealed by RTD as well. In Figure \ref{fig:gnn_converg_2}, we take a closer look at evolution of RTD-Lite vs. RTD throughout training. In most cases, RTD and RTD-Lite are well aligned. Comparison of SAGE layers' representations (Figure \ref{fig:gnn_converg_2_sage}) provides the only case when RTD-Lite behaves differently. To some extent, this is natural because RTD-Lite takes into account only $0$-dimensional topological features while RTD considers $1$-dimensional topological features as well.  

\vskip-.1in

\subsection{Adversarial attacks analysis}
\label{app:gnn_adversarial}

\begin{figure*}[tbp]
\centering
\begin{subfigure}{0.235\textwidth}
\includegraphics[width=\textwidth]{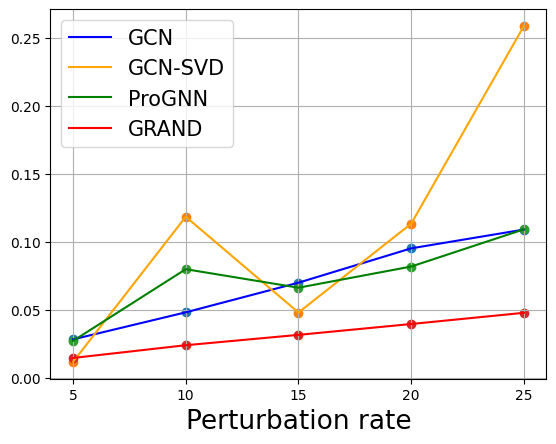}
\caption{1-CKA, GE}
\label{fig:gnn_adv_cka_ge}
\end{subfigure}
\begin{subfigure}{0.24\textwidth}
\includegraphics[width=\textwidth]{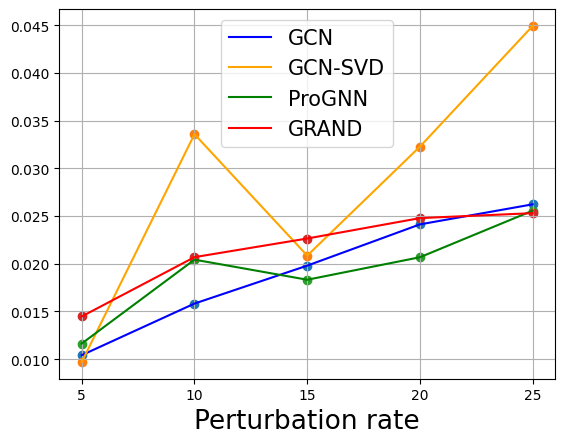}
\caption{NSA, GE}
\label{fig:gnn_adv_nsa_ge}
\end{subfigure}
\begin{subfigure}{0.23\textwidth}
\includegraphics[width=\textwidth]{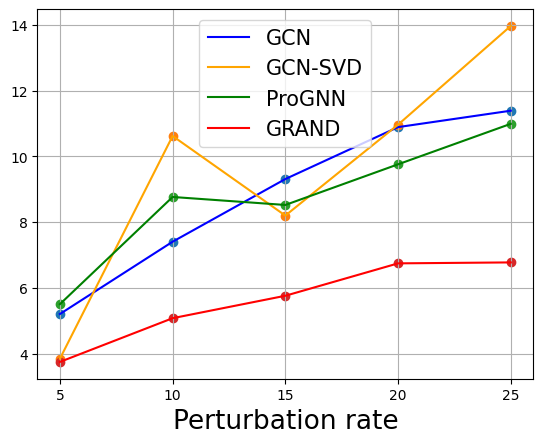}
\caption{RTD, GE}
\label{fig:gnn_adv_rtd_ge}
\end{subfigure}
\begin{subfigure}{0.23\textwidth}
\includegraphics[width=\textwidth]{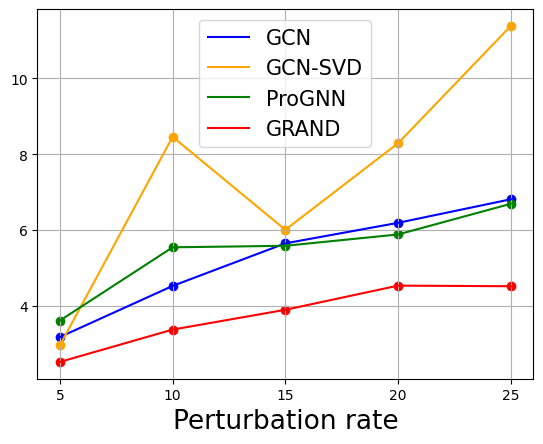}
\caption{RTD-Lite, GE}
\label{fig:gnn_adv_rtd_lite_ge}
\end{subfigure}
\begin{subfigure}{0.235\textwidth}
\includegraphics[width=\textwidth]{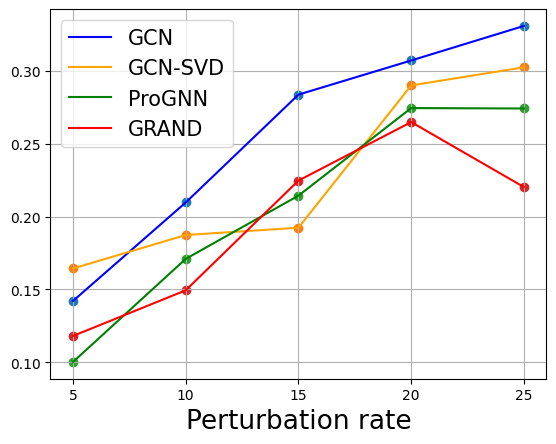}
\caption{1-CKA, GP}
\label{fig:gnn_adv_cka_gp}
\end{subfigure}
\begin{subfigure}{0.24\textwidth}
\includegraphics[width=\textwidth]{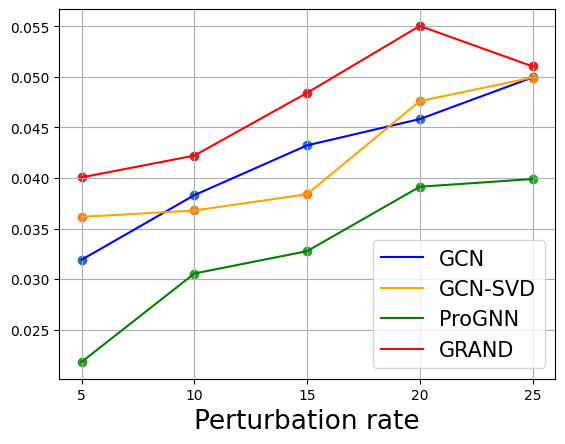}
\caption{NSA, GP}
\label{fig:gnn_adv_nsa_gp}
\end{subfigure}
\begin{subfigure}{0.23\textwidth}
\includegraphics[width=\textwidth]{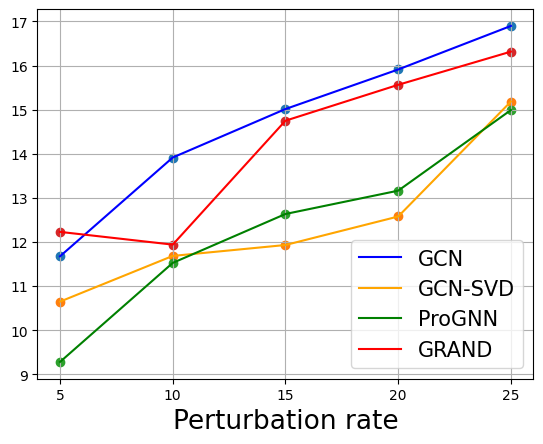}
\caption{RTD, GP}
\label{fig:gnn_adv_rtd_gp}
\end{subfigure}
\begin{subfigure}{0.23\textwidth}
\includegraphics[width=\textwidth]{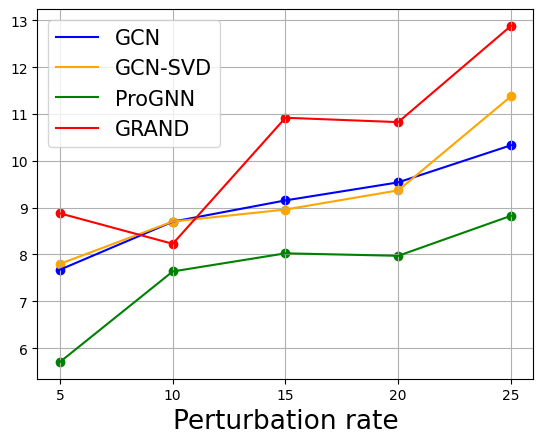}
\caption{RTD-Lite, GP}
\label{fig:gnn_adv_rtd_lite_gp}
\end{subfigure}
\begin{subfigure}{0.24\textwidth}
\includegraphics[width=\textwidth]{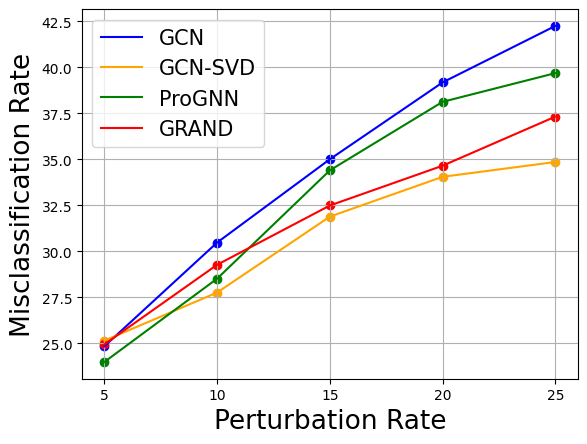}
\caption{Misclass. rate, GE}
\label{fig:gnn_adv_misclass_ge}
\end{subfigure}
\begin{subfigure}{0.24\textwidth}
\includegraphics[width=\textwidth]{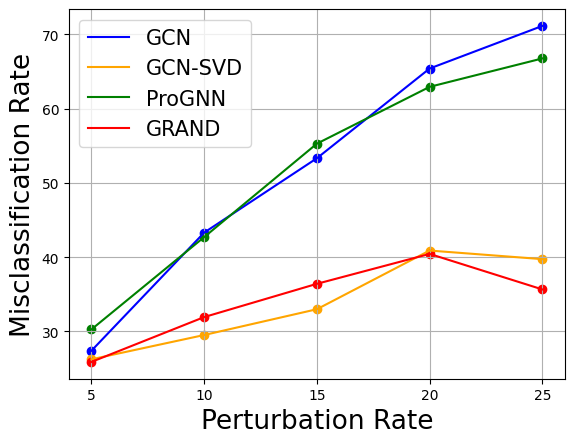}
\caption{Misclass. rate, GP}
\label{fig:gnn_adv_misclass_gp}
\end{subfigure}
\begin{subfigure}{0.235\textwidth}
\includegraphics[width=\textwidth]{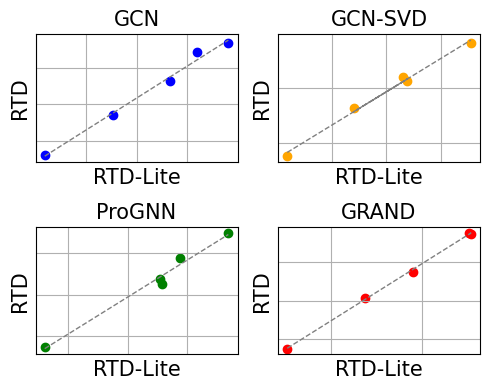}
\caption{RTD-Lite vs. RTD, GE}
\label{fig:gnn_adv_align_ge}
\end{subfigure}
\begin{subfigure}{0.235\textwidth}
\includegraphics[width=\textwidth]{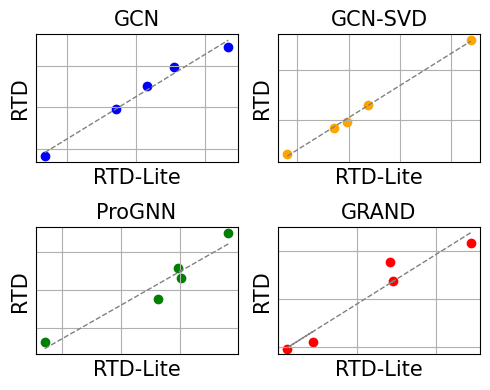}
\caption{RTD-Lite vs. RTD, GP}
\label{fig:gnn_adv_align_gp}
\end{subfigure}
\caption{Dissimilarity in GNNs' representations under adversarial attacks of various intensity as measured by CKA, NSA, RTD, RTD-Lite in comparison with misclassification rate: (a - d, i, k) - global evasion attack (GE), (e - h, j, l) - global poisoning attack (GP).}
\label{fig:gnn_adversarial}
\end{figure*}

The work \citet{ebadulla2024normalized} studies whether dissimilarity between GNNs' representations under adversarial attacks of various intensity is aligned with misclassification rate. We further extend this evaluation and verify whether RTD-Lite is able to capture deterioration of GNNs' classification performance. Following the experimental setup from loc.cit.,
we apply global evasion and global poisoning adversarial attacks to GCN \citet{} and robust GNN variants - GCN-SVD \citep{entezari2020all}, ProGNN \citep{jin2020graph}, GRAND \citep{chamberlain2021grand} - under various intensity rates. In this setup, adversarial attacks are constructed by introducing perturbations in the range $5\% - 25\%$ into adjacency matrices. For each value of perturbation rate, we compute the dissimilarity between the initial representations and representations under attack. Figure \ref{fig:gnn_adversarial} demonstrates that, in general, RTD-lite captures the same tendency as misclassification rate, similar to other metrics. Our experiments also confirm the distinctive behavior of all the representation dissimilarity metrics in the case of evasion attacks for GCN-SVD as was mentioned in \citet{ebadulla2024normalized}. %
This effect was attributed to vulnerability in the structure that was also noticed in \citet{mujkanovic2022are}. Therefore, RTD-Lite and other metrics are able to reflect potential weaknesses that may be not evident when analysing the misclassification rate solely. In Figures \ref{fig:gnn_adv_align_ge}, \ref{fig:gnn_adv_align_gp}, we demonstrate that RTD-Lite scores are well aligned with RTD values. Hence, RTD-Lite can be used as an approximation of RTD when its intensive computation is not feasible.

\subsection{Additional Evaluation for the experiments from Section \ref{diff_init}}

\begin{figure*}[t]
\centering
\includegraphics[width=\textwidth]{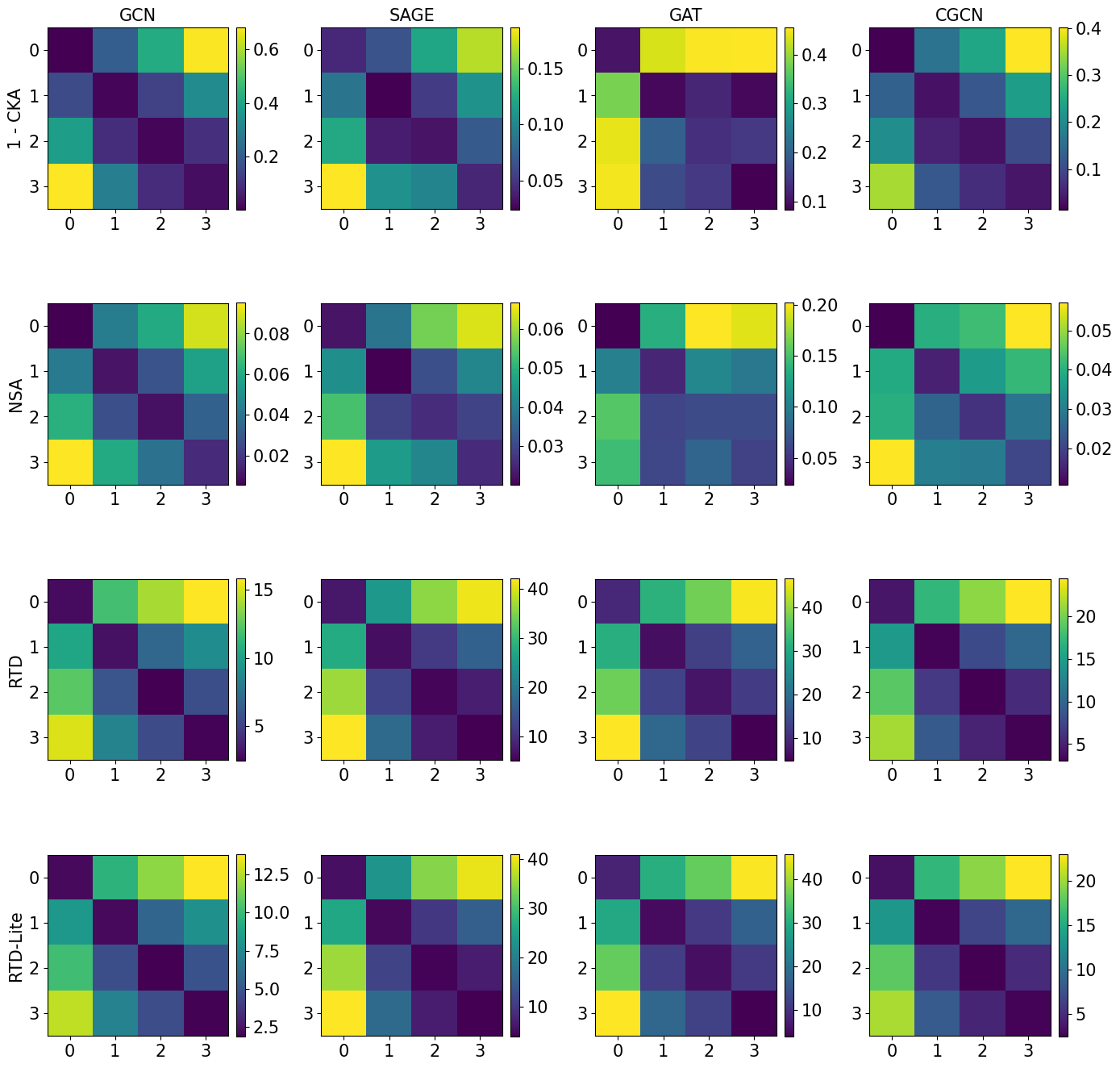}
\caption{Additional evaluation of CKA, NSA, RTD for the experiment from Section \ref{diff_init}.}
\label{fig:diff_init_additional}
\end{figure*}

In this section, we provide additional evaluation of CKA, NSA, RTD representation dissimilarity measures for GNN models. Identical to setup from Section \ref{diff_init}, we compare layers' representations from the GNN models (GCN, GraphSAGE, GAT, ClusterGCN) trained with different initialization. Figure \ref{fig:diff_init_additional} provides the mentioned evaluation.

\subsection{Training details on GNN experiments}
\label{app:gnn_training}

For the experiments on measuring the dissimilarity in representations of GNN, we followed the setup from \citet{ebadulla2024normalized}.
In our experiments with adversarial attacks analysis from Section \ref{app:gnn_adversarial}, we trained GCN, GCN-SVD, ProGNN, GRAND also following  the setup from loc.cit.
For the experiments from Sections \ref{diff_init}, \ref{app:gnn_convergence}, we trained GCN, SAGE, GAT, CGCN models with hyperparameters listed in loc.cit.,
Table $6$. Table \ref{app:tbl:gnn_acc} provides accuracy of our obtained models. We performed the experiments using one Nvidia Titan RTX GPU.

\addtolength{\tabcolsep}{-3pt}
\begin{table}
  \caption{GNN test accuracy on Amazon Computer Dataset.}
  \label{app:tbl:gnn_acc}
  \centering
  \begin{tabular}{lc}
    \toprule
    Architecture & Accuracy \\
    \midrule
    GCN & $0.8677 \pm 0.0004$\\
    SAGE & $0.9124 \pm 0.0007$\\
    GAT & $0.8920 \pm 0.0010$\\
    CGCN & $0.8689 \pm 0.0011$\\
    \bottomrule
  \end{tabular}
\end{table}
\addtolength{\tabcolsep}{2pt}

\section{Experiment details for gradient optimization task}

In all reported experiments, for all autoencoder methods (RTD-Lite Autoencoder, RTD-Autoencoder, TopoAE and vanilla Autoencoder) we used the same architecture of deep autoencoder with 3 fully-connected hidden layers having $512, 256, \text{and } 128$ neurons each and ReLU activation function. We used Adam optimizer with learning rate of $0.0001$. We did not perform  extensive search for optimal value of hyperparameter $\lambda$ governing the ratio between reconstruction and RTD-Lite loss function components and used $\lambda = 1.0$ (except for COIL-20 dataset, where $\lambda=5.0$ was used).

For the \textit{Spheres} dataset, training was performed for 100 epochs; RTD-AE was trained for the first 10 epochs using only the reconstruction loss, as suggested in the original paper. The batch size was set to $80$.

For the F-MNIST and COIL-20 datasets, training was performed for 250 epochs; RTD-AE was trained for the first 60 epochs using only the reconstruction loss, as suggested in the original paper. The batch size was set to $256$.

\section{Applications at Large Language Models}

RTD-Lite can be used to study changes in model internal data representations as caused by the fine-tuning process. For instance, its high performance allows for applications with modern Large Language Models with large context spans.
We performed experiments with LLaMA2-7B model, using RTD-Lite to compare topology of intermediate embeddings between base model and its version additionally fine-tuned at processing dialogues. 

For this experiment, we gathered two small corpora of English texts and dialogues (1000 entries in each), where texts on average contain $320 \pm 52$ tokens, dialogues --- $181 \pm 23$ (this translates into number of vertices in corresponding graphs). 

\begin{figure*}[h]
\centering
\includegraphics[width=0.9\textwidth]{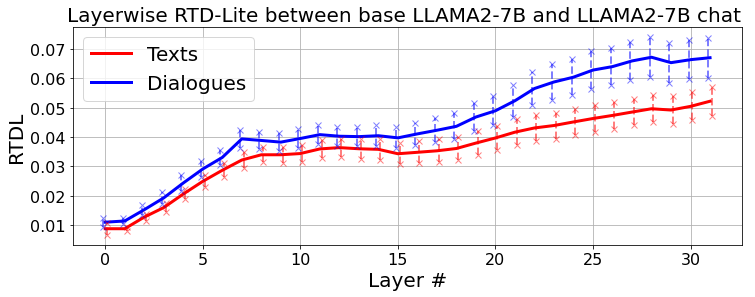}
\caption{Distributions of RTDL between embeddings of textual and dialogic data from corresponding layers of LLaMA2-7B-base and -chat models. Solid line mark the average, vertical lines -- standard deviation.}
\label{fig:llamas_appnd}
\end{figure*}

\paragraph{Comparing intermediate embeddings}
 We passed each entry through base and chat-tuned models and calculated RTDL between embeddings from each of 32 pairs of corresponding layers. 
In total, for this experiment we calculated RTDL for $2 * 1,000 * 32 = 64,000$ pairs of graphs originating from Euclidean distances between points in $\mathrm{R}^{4096}$, and on average, processing each took $\approx 0.07$ sec for texts and $\approx 0.02$ sec for dialogues.
On Figure \ref{fig:llamas_appnd}  the distributions of RTDL for textual and dialogic data respectfully are plotted. We can see that in the lower half of the model (up to layer 15) difference between RTDL of text and dialogues is small and quite stable, but it begins to grow for upper layers of the model, and from layer 24 upwards it exceeds two standard deviations. 

\paragraph{Comparing attention maps}

We performed similar experiments to compare topology of the corresponding attention maps from base and chat-tuned models. Attention maps can be viewed as weighted directed graphs with vertices corresponding to text tokens. LLaMA2 implements triangular attention (it calculates attention weights only between a token and all previous), therefore its attention maps can be transformed into undirected graphs without loss of information. Unlike experiments from the previous paragraph, these graphs do not have realization as distance matrices of a point cloud. 

LLaMA2-7B has 1,024 attention heads (32 in each layer), thus we calculated RTDL for $2 * 1,000 * 1,024 = 2,048,000$ pairs of graphs.

\begin{figure*}[h]
\centering
\begin{subfigure}{0.495\textwidth}
\includegraphics[width=\textwidth]{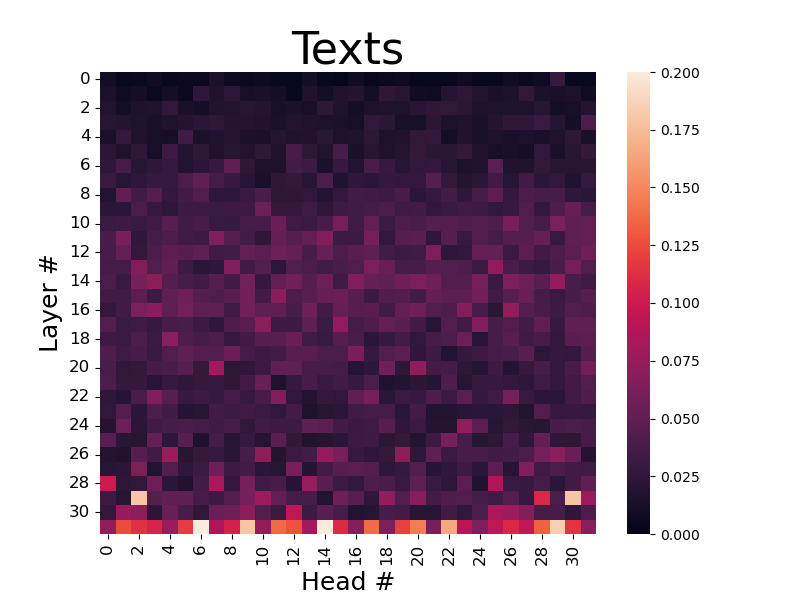} 
\end{subfigure}
\begin{subfigure}{0.495\textwidth}
\includegraphics[width=\textwidth]{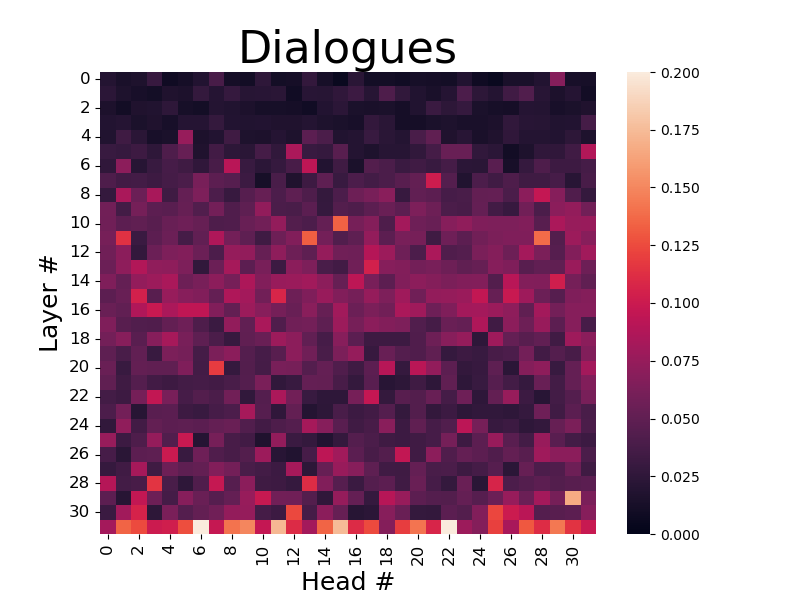} 
\end{subfigure}
\caption{Mean values of RTDL between same heads of LLaMA2-7B-base and -chat-tuned models for same inputs. On the left heatmap  for textual data, on the right --- for dialogic.}
\label{fig:llamas_heads_appnd}
\end{figure*}

Figure \ref{fig:llamas_heads_appnd} presents the results.

These observations support the idea that during fine-tuning of a deep model only weights from its later layers are subject to any significant changes. We hope that this approach can be used to better identify border between layers that do and do not require fine-tuning, thus lowering computations needed for model fine-tuning.

\section{RTD-Lite barcode and the dimensionality of $\ker{(r_{0, \alpha})}$}
\label{app:proof}

\begin{lemma}
Dimensionality of $\ker{(r_{0, \alpha})}$ equals to the number of intervals in RTD-Lite barcode that contains \nolinebreak $\alpha$.
\end{lemma}

\begin{proof}

By the rank-nullity theorem for the map 

\[
r_{0,\alpha}: H_0(A^{\leq\alpha}) \rightarrow H_0(C^{\leq\alpha}),
\]

we obtain: 

\begin{equation}
    \dim{H_0(A^{\leq\alpha})} = \dim{\ker{(r_{0, \alpha})}} + \dim{\mathrm{Im}(r_0,\alpha)}.
    \label{app:eq1}
\end{equation}

By construction of the graph $C$, $r_{0,\alpha}$ is epimorphism, i.e. $\mathrm{Im}(r_0,\alpha) = H_0(C^{\leq\alpha})$. Thus, Equation \ref{app:eq1} is equivalent to:

\begin{equation}
    \dim{\ker{(r_{0, \alpha})}} = \dim{H_0(A^{\leq\alpha})} - \dim{H_0(C^{\leq\alpha})}.
    \label{app:eq2}
\end{equation}

Let us see how $\dim{\ker{(r_{0, \alpha})}}$ changes when increasing the value of $\alpha$ via analysing the dimensionalities of homology groups $H_0(A^{\leq\alpha}), H_0(C^{\leq\alpha})$. 
First, note that for $\alpha = 0$, both $A^{\leq\alpha}$ and $C^{\leq\alpha}$ consist only of $n$ vertices with no edges. Thus, $\dim{H_0(A^{\leq\alpha})} = \dim{H_0(C^{\leq\alpha})} = n$ and $\dim{\ker{(r_{0, \alpha})}} = 0$.
Also, the intersection of the RTD-Lite barcode with $\alpha=0$ line is empty.

Now let's add to the set of vertices the edges of $C\supseteq A$  one by one in the increasing weight order.  Let us increase the value of $\alpha$ from $\alpha_1$ to the next weight $\alpha_2$ of an edge from $C$, $\alpha_2 > \alpha_1$. First, consider the case when the new edge $e$ appends to $A^{\leq\alpha_2}$ compared to $A^{\leq\alpha_1}$. If $e$ connects the two vertices from the same connected component, then $\dim{H_0(A^{\leq\alpha_2})} = \dim{H_0(A^{\leq\alpha_1})}$ and $\dim{H_0(C^{\leq\alpha_2})} = \dim{H_0(C^{\leq\alpha_1})}$ and $\dim{\ker{(r_{0, \alpha_2})}} = \dim{\ker{(r_{0, \alpha_1})}}$. If $e$ connects the vertices from the two different connected components, then the two components merge together and $\dim{H_0(A^{\leq\alpha_2})} = \dim{H_0(A^{\leq\alpha_1})} - 1$. If $e$ corresponds to the vertices from the same connected component in $C^{\leq\alpha_1}$, then $\dim{H_0(C^{\leq\alpha_2})} = \dim{H_0(C^{\leq\alpha_1})}$. In this case, $\dim{\ker{(r_{0, \alpha_2})}} = \dim{H_0(A^{\leq\alpha_2})} - \dim{H_0(C^{\leq\alpha_2})} = (\dim{H_0(A^{\leq\alpha_1})} - 1) - \dim{H_0(C^{\leq\alpha_1})} = \dim{\ker{(r_{0, \alpha_1})}} - 1$. If $e$ corresponds to the vertices from different connected components, then $\dim{H_0(C^{\leq\alpha_2})} = \dim{H_0(C^{\leq\alpha_1})} - 1$ and $\dim{\ker{(r_{0, \alpha_2})}} = \dim{H_0(A^{\leq\alpha_2})} - \dim{H_0(C^{\leq\alpha_2})} = (\dim{H_0(A^{\leq\alpha_1})} - 1) - (\dim{H_0(C^{\leq\alpha_1})} - 1) = \dim{\ker{(r_{0, \alpha_1})}}$. 

Second, consider the case when $A^{\leq\alpha_2} = A^{\leq\alpha_1}$ (and $H_0(A^{\leq\alpha_2}) = H_0(A^{\leq\alpha_1})$) but a new edge $e$ appears in $C^{\leq\alpha_2}$ compared to $C^{\leq\alpha_1}$. If $e$ corresponds to the vertices from the same connected component, then $H_0(C^{\leq\alpha_2}) = H_0(C^{\leq\alpha_1})$ and $\dim{\ker{(r_{0, \alpha_2})}} = \dim{\ker{(r_{0, \alpha_1})}}$. If $e$ corresponds to the vertices from different connected components, then $H_0(C^{\leq\alpha_2}) = H_0(C^{\leq\alpha_1}) - 1$ and $\dim{\ker{(r_{0, \alpha_2})}} = \dim{H_0(A^{\leq\alpha_2})} - \dim{H_0(C^{\leq\alpha_2})} = 
\dim{H_0(A^{\leq\alpha_1})} - (\dim{H_0(C^{\leq\alpha_1})} - 1) = \dim{\ker{(r_{0, \alpha_1})}} + 1$. 

Thus, there are two cases of a change in $\dim{\ker{(r_{0, \alpha_1})}}$:
\begin{itemize}
    \item $\dim{\ker{(r_{0, \alpha_2})}} = \dim{\ker{(r_{0, \alpha_1})}} - 1$.
    In this case, the two connected components that were merged in $C^{\leq\alpha}$ for some $\alpha < \alpha_2$ now merge in $A^{\leq\alpha_2}$. This corresponds to closing of the interval $(\alpha, \alpha_2)$ in RTD-Lite barcode.
    \item $\dim{\ker{(r_{0, \alpha_2})}} = \dim{\ker{(r_{0, \alpha_1})}} + 1$.
    In this case, there are two connected components that merge in $C^{\leq\alpha_2}$ and not (yet) merge in $A^{\leq\alpha_2}$. This corresponds to opening of the interval $(\alpha_2, \alpha)$ in RTD-Lite barcode for some $\alpha > \alpha_2$.
\end{itemize}

Therefore, the decrease and increase in \(\dim \ker (r_0, \alpha)\) correspond to the closing and opening of intervals in the RTD-Lite barcode. Thus, for each \(\alpha\), \(\dim \ker (r_0, \alpha)\) equals the number of intervals in the RTD-Lite barcode that contain \(\alpha\).
\end{proof}

\section{Licenses of the used assets}
\label{sec:licenses}

Here is the list of licenses for the assets (models and datasets) that we used in this work.

\begin{itemize}
    \item CIFAR-10, Fashion MNIST (F-MNIST): \underline{MIT license}
    \item COIL-20: \underline{GNU General Public License v2.0}
    \item MNIST: \underline{GNU General Public License v3.0}
    \item LLaMA2-7B: \underline{LLAMA 2 COMMUNITY LICENSE AGREEMENT}
    
\end{itemize}

We do not publish any new models or datasets with this paper.

\end{document}

\onecolumn
\aistatstitle{AISTATS 2025 Instruction: \\
Supplementary Materials}

\section{FORMATTING INSTRUCTIONS}

To prepare a supplementary pdf file, we ask the authors to use \texttt{aistats2025.sty} as a style file and to follow the same formatting instructions as in the main paper.
The only difference is that the supplementary material must be in a \emph{single-column} format.
You can use \texttt{supplement.tex} in our starter pack as a starting point, or append the supplementary content to the main paper and split the final PDF into two separate files.

Note that reviewers are under no obligation to examine your supplementary material.

\section{MISSING PROOFS}

The supplementary materials may contain detailed proofs of the results that are missing in the main paper.

\subsection{Proof of Lemma 3}

\textit{In this section, we present the detailed proof of Lemma 3 and then [ ... ]}

\section{ADDITIONAL EXPERIMENTS}

If you have additional experimental results, you may include them in the supplementary materials.

\subsection{The Effect of Regularization Parameter}

\textit{Our algorithm depends on the regularization parameter $\lambda$. Figure 1 below illustrates the effect of this parameter on the performance of our algorithm. As we can see, [ ... ]}

\vfill

%% file: math_commands.tex
\usepackage{amsmath,amsfonts,bm}

\def\eqref#1{equation~\ref{#1}}

\def\1{\bm{1}}

\DeclareMathAlphabet{\mathsfit}{\encodingdefault}{\sfdefault}{m}{sl}
\SetMathAlphabet{\mathsfit}{bold}{\encodingdefault}{\sfdefault}{bx}{n}

%% file: main.bbl
\begin{thebibliography}{40}
\providecommand{\natexlab}[1]{#1}
\providecommand{\url}[1]{\texttt{#1}}
\expandafter\ifx\csname urlstyle\endcsname\relax
  \providecommand{\doi}[1]{doi: #1}\else
  \providecommand{\doi}{doi: \begingroup \urlstyle{rm}\Url}\fi

\bibitem[Barannikov(1994)]{barannikov1994framed}
S.~Barannikov.
\newblock The framed {M}orse complex and its invariants.
\newblock \emph{Advances in Soviet Mathematics}, 21:\penalty0 93--116, 1994.

\bibitem[Barannikov(2021)]{barannikov2021canonical}
S.~Barannikov.
\newblock Canonical forms $= $ persistence diagrams. tutorial.
\newblock In \emph{EuroCG2021 The 37th European Workshop on Computational Geometry}, 2021.

\bibitem[Barannikov et~al.(2021)Barannikov, Trofimov, Balabin, and Burnaev]{barannikov2021representation}
S.~Barannikov, I.~Trofimov, N.~Balabin, and E.~Burnaev.
\newblock Representation topology divergence: A method for comparing neural network representations.
\newblock In \emph{Proceedings of the 39th International Conference on Machine Learning}, volume 162, pages 1607--1626, 2021.

\bibitem[Borgwardt and Kriegel(2005)]{Borgwardt2005ShortestPathKernel}
K.~M. Borgwardt and H.-P. Kriegel.
\newblock Shortest-path kernels on graphs.
\newblock In \emph{Proceedings of the 5th IEEE International Conference on Data Mining (ICDM)}, pages 74--81. IEEE, 2005.

\bibitem[Carri{\`e}re et~al.(2021)Carri{\`e}re, Cuturi, and Oudot]{carriere2021optimizing}
M.~Carri{\`e}re, M.~Cuturi, and S.~Oudot.
\newblock Optimizing persistent homology based functions.
\newblock \emph{International Conference on Machine Learning (ICML)}, pages 1294--1303, 2021.

\bibitem[Chamberlain et~al.(2021)Chamberlain, Rowbottom, Gorinova, Webb, Rossi, and Bronstein]{chamberlain2021grand}
B.~P. Chamberlain, J.~Rowbottom, M.~I. Gorinova, S.~D. Webb, E.~Rossi, and M.~M. Bronstein.
\newblock {GRAND}: Graph neural diffusion.
\newblock In \emph{The Symbiosis of Deep Learning and Differential Equations}, 2021.
\newblock URL \url{https://openreview.net/forum?id=_1fu_cjsaRE}.

\bibitem[Chazal and Michel(2021)]{chazal2021introduction}
F.~Chazal and B.~Michel.
\newblock An introduction to topological data analysis: fundamental and practical aspects for data scientists.
\newblock \emph{Frontiers in artificial intelligence}, 4:\penalty0 667963, 2021.

\bibitem[Chen et~al.(2021)Chen, Coskunuzer, and Gel]{chen2021topological}
Y.~Chen, B.~Coskunuzer, and Y.~Gel.
\newblock Topological relational learning on graphs.
\newblock \emph{Advances in neural information processing systems}, 34:\penalty0 27029--27042, 2021.

\bibitem[Chiang et~al.(2019)Chiang, Liu, Si, Li, Bengio, and Hsieh]{chiang2019cluster}
W.-L. Chiang, X.~Liu, S.~Si, Y.~Li, S.~Bengio, and C.-J. Hsieh.
\newblock Cluster-gcn: An efficient algorithm for training deep and large graph convolutional networks.
\newblock In \emph{Proceedings of the 25th ACM SIGKDD international conference on knowledge discovery \& data mining}, pages 257--266, 2019.

\bibitem[Damrich and Hamprecht(2021)]{damrich2021umap}
S.~Damrich and F.~A. Hamprecht.
\newblock On umap's true loss function.
\newblock \emph{Advances in Neural Information Processing Systems}, 34:\penalty0 5798--5809, 2021.

\bibitem[Ebadulla and Singh(2024)]{ebadulla2024normalized}
D.~Ebadulla and A.~Singh.
\newblock Normalized space alignment: A versatile metric for representation space discrepancy minimization, 2024.
\newblock URL \url{https://openreview.net/forum?id=5HGPR6fg2S}.

\bibitem[Entezari et~al.(2020)Entezari, Al-Sayouri, Darvishzadeh, and Papalexakis]{entezari2020all}
N.~Entezari, S.~A. Al-Sayouri, A.~Darvishzadeh, and E.~E. Papalexakis.
\newblock All you need is low (rank): Defending against adversarial attacks on graphs.
\newblock In \emph{Proceedings of the 13th International Conference on Web Search and Data Mining}, WSDM '20, page 169–177, New York, NY, USA, 2020. Association for Computing Machinery.
\newblock ISBN 9781450368223.
\newblock \doi{10.1145/3336191.3371789}.
\newblock URL \url{https://doi.org/10.1145/3336191.3371789}.

\bibitem[Hamilton et~al.(2017)Hamilton, Ying, and Leskovec]{hamilton2017inductive}
W.~Hamilton, Z.~Ying, and J.~Leskovec.
\newblock Inductive representation learning on large graphs.
\newblock \emph{Advances in neural information processing systems}, 30, 2017.

\bibitem[Jin et~al.(2020)Jin, Ma, Liu, Tang, Wang, and Tang]{jin2020graph}
W.~Jin, Y.~Ma, X.~Liu, X.~Tang, S.~Wang, and J.~Tang.
\newblock Graph structure learning for robust graph neural networks.
\newblock In \emph{Proceedings of the 26th ACM SIGKDD International Conference on Knowledge Discovery \& Data Mining}, KDD '20, page 66–74, New York, NY, USA, 2020. Association for Computing Machinery.
\newblock ISBN 9781450379984.
\newblock \doi{10.1145/3394486.3403049}.
\newblock URL \url{https://doi.org/10.1145/3394486.3403049}.

\bibitem[Kim et~al.(2020)Kim, Lee, Kang, and Choo]{kim2020pllay}
H.~Kim, J.~Lee, U.~Kang, and J.~Choo.
\newblock Pl-lay: Efficient layer-wise topological regularization.
\newblock \emph{Proceedings of the 37th International Conference on Machine Learning (ICML)}, 2020.

\bibitem[Kipf and Welling(2016)]{kipf2016semi}
T.~N. Kipf and M.~Welling.
\newblock Semi-supervised classification with graph convolutional networks.
\newblock \emph{arXiv preprint arXiv:1609.02907}, 2016.

\bibitem[Kong et~al.(2024)Kong, Zhang, and Li]{kong2024learning}
D.~Kong, A.~Zhang, and Y.~Li.
\newblock Learning persistent community structures in dynamic networks via topological data analysis.
\newblock In \emph{Proceedings of the AAAI Conference on Artificial Intelligence}, volume~38, pages 8617--8626, 2024.

\bibitem[Kornblith et~al.(2019)Kornblith, Norouzi, Lee, and Hinton]{kornblith2019similarity}
S.~Kornblith, M.~Norouzi, H.~Lee, and G.~Hinton.
\newblock Similarity of neural network representations revisited.
\newblock In \emph{International Conference on Machine Learning}, pages 3519--3529. PMLR, 2019.

\bibitem[Leygonie et~al.(2021)Leygonie, Rieck, Moor, and Borgwardt]{leygonie2021framework}
J.~Leygonie, B.~Rieck, M.~Moor, and K.~Borgwardt.
\newblock A framework for differentiable topological layer based on persistent homology.
\newblock \emph{Conference on Neural Information Processing Systems (NeurIPS)}, 2021.

\bibitem[Luo et~al.(2021)Luo, You, Guo, and Ji]{luo2021topology}
D.~Luo, H.~You, Y.~Guo, and S.~Ji.
\newblock Topology-preserving deep learning for graph representations.
\newblock \emph{Proceedings of the 29th ACM International Conference on Multimedia}, pages 3734--3742, 2021.

\bibitem[McInnes et~al.(2018)McInnes, Healy, and Melville]{mcinnes2018umap}
L.~McInnes, J.~Healy, and J.~Melville.
\newblock Umap: Uniform manifold approximation and projection for dimension reduction.
\newblock \emph{arXiv preprint arXiv:1802.03426}, 2018.

\bibitem[Moor et~al.(2020)Moor, Horn, Rieck, and Borgwardt]{moor2020topological}
M.~Moor, M.~Horn, B.~Rieck, and K.~Borgwardt.
\newblock Topological autoencoders.
\newblock In \emph{International conference on machine learning}, pages 7045--7054. PMLR, 2020.

\bibitem[Mujkanovic et~al.(2022)Mujkanovic, Geisler, G\"{u}nnemann, and Bojchevski]{mujkanovic2022are}
F.~Mujkanovic, S.~Geisler, S.~G\"{u}nnemann, and A.~Bojchevski.
\newblock Are defenses for graph neural networks robust?
\newblock In S.~Koyejo, S.~Mohamed, A.~Agarwal, D.~Belgrave, K.~Cho, and A.~Oh, editors, \emph{Advances in Neural Information Processing Systems}, volume~35, pages 8954--8968. Curran Associates, Inc., 2022.

\bibitem[Nene et~al.(1996)Nene, Nayar, Murase, et~al.]{nene1996columbia}
S.~A. Nene, S.~K. Nayar, H.~Murase, et~al.
\newblock Columbia object image library (coil-20).
\newblock 1996.

\bibitem[Qin et~al.(2024)Qin, Fasy, Wenk, and Summa]{qin2024rapid}
Y.~Qin, B.~T. Fasy, C.~Wenk, and B.~Summa.
\newblock Rapid and precise topological comparison with merge tree neural networks.
\newblock \emph{arXiv preprint arXiv:2404.05879}, 2024.

\bibitem[Raghu et~al.(2017)Raghu, Gilmer, Yosinski, and Sohl-Dickstein]{raghu2017svcca}
M.~Raghu, J.~Gilmer, J.~Yosinski, and J.~Sohl-Dickstein.
\newblock Svcca: Singular vector canonical correlation analysis for deep learning dynamics and interpretability.
\newblock \emph{arXiv preprint arXiv:1706.05806}, 2017.

\bibitem[Sanfeliu and Fu(1983)]{Sanfeliu1983GraphDistance}
A.~Sanfeliu and K.-S. Fu.
\newblock A distance measure between attributed relational graphs for pattern recognition.
\newblock \emph{IEEE Transactions on Systems, Man, and Cybernetics}, 13\penalty0 (3):\penalty0 353--362, 1983.

\bibitem[Shchur et~al.(2018)Shchur, Mumme, Bojchevski, and G{\"u}nnemann]{shchur2018pitfalls}
O.~Shchur, M.~Mumme, A.~Bojchevski, and S.~G{\"u}nnemann.
\newblock Pitfalls of graph neural network evaluation.
\newblock \emph{arXiv preprint arXiv:1811.05868}, 2018.

\bibitem[Shervashidze et~al.(2011)Shervashidze, Schweitzer, van Leeuwen, Mehlhorn, and Borgwardt]{Shervashidze2011WeisfeilerLehman}
N.~Shervashidze, P.~Schweitzer, E.~J. van Leeuwen, K.~Mehlhorn, and K.~M. Borgwardt.
\newblock Weisfeiler-lehman graph kernels.
\newblock \emph{J. Mach. Learn. Res.}, 12:\penalty0 2539--2561, 2011.

\bibitem[Springenberg et~al.(2014)Springenberg, Dosovitskiy, Brox, and Riedmiller]{springenberg2014striving}
J.~T. Springenberg, A.~Dosovitskiy, T.~Brox, and M.~Riedmiller.
\newblock Striving for simplicity: The all convolutional net.
\newblock \emph{arXiv preprint arXiv:1412.6806}, 2014.

\bibitem[Trofimov et~al.(2023)Trofimov, Cherniavskii, Tulchinskii, Balabin, Burnaev, and Barannikov]{trofimov2023learning}
I.~Trofimov, D.~Cherniavskii, E.~Tulchinskii, N.~Balabin, E.~Burnaev, and S.~Barannikov.
\newblock Learning topology-preserving data representations.
\newblock In \emph{ICLR 2023 International Conference on Learning Representations}, volume 2023, 2023.

\bibitem[Tsitsulin et~al.(2020)Tsitsulin, Munkhoeva, Mottin, Karras, Bronstein, Oseledets, and Mueller]{tsitsulin2019shape}
A.~Tsitsulin, M.~Munkhoeva, D.~Mottin, P.~Karras, A.~Bronstein, I.~Oseledets, and E.~Mueller.
\newblock The shape of data: Intrinsic distance for data distributions.
\newblock In \emph{International Conference on Learning Representations}, 2020.

\bibitem[Velickovic et~al.(2017)Velickovic, Cucurull, Casanova, Romero, Lio, Bengio, et~al.]{velickovic2017graph}
P.~Velickovic, G.~Cucurull, A.~Casanova, A.~Romero, P.~Lio, Y.~Bengio, et~al.
\newblock Graph attention networks.
\newblock \emph{stat}, 1050\penalty0 (20):\penalty0 10--48550, 2017.

\bibitem[Vishwanathan et~al.(2010)Vishwanathan, Schraudolph, Kondor, and Borgwardt]{vishwanathan10a}
S.~Vishwanathan, N.~N. Schraudolph, R.~Kondor, and K.~M. Borgwardt.
\newblock Graph kernels.
\newblock \emph{Journal of Machine Learning Research}, 11\penalty0 (40):\penalty0 1201--1242, 2010.

\bibitem[Wang et~al.(2021)Wang, Huang, Rudin, and Shaposhnik]{wang2021understanding}
Y.~Wang, H.~Huang, C.~Rudin, and Y.~Shaposhnik.
\newblock Understanding how dimension reduction tools work: an empirical approach to deciphering t-{SNE}, {UMAP}, {TriMAP}, and {PaCMAP} for data visualization.
\newblock \emph{J Mach. Learn. Res}, 22:\penalty0 1--73, 2021.

\bibitem[Wen et~al.(2024)Wen, Chen, and Chen]{wen2024tensor}
T.~Wen, E.~Chen, and Y.~Chen.
\newblock Tensor-view topological graph neural network.
\newblock In \emph{International Conference on Artificial Intelligence and Statistics}, pages 4330--4338. PMLR, 2024.

\bibitem[Xiao et~al.(2017)Xiao, Rasul, and Vollgraf]{xiao2017online}
H.~Xiao, K.~Rasul, and R.~Vollgraf.
\newblock Fashion-mnist: a novel image dataset for benchmarking machine learning algorithms.
\newblock \emph{ArXiv}, abs/1708.07747, 2017.
\newblock URL \url{https://api.semanticscholar.org/CorpusID:702279}.

\bibitem[Xu et~al.(1990)Xu, Olman, and Xu]{Xu1990ParallelMSTClustering}
J.~Xu, V.~Olman, and D.~Xu.
\newblock A fast parallel algorithm for clustering using minimum spanning trees.
\newblock \emph{IEEE Transactions on Parallel and Distributed Systems}, 1\penalty0 (3):\penalty0 355--365, 1990.

\bibitem[Zahn(1971)]{Zahn1971GestaltClusters}
C.~T. Zahn.
\newblock Graph-theoretical methods for detecting and describing gestalt clusters.
\newblock \emph{IEEE Transactions on Computers}, C-20\penalty0 (1):\penalty0 68--86, 1971.

\bibitem[Zomorodian(2001)]{zomorodian2001computing}
A.~J. Zomorodian.
\newblock \emph{Computing and comprehending topology: Persistence and hierarchical {M}orse complexes. PhD Thesis}.
\newblock University of Illinois at Urbana-Champaign, 2001.

\end{thebibliography}
